\documentclass{article}
\usepackage{amssymb}
\usepackage{graphicx}
\usepackage{wrapfig}
\usepackage[table]{xcolor}
\usepackage{amsmath}
\usepackage{color, soul} 
\usepackage[font=small]{caption}  
\usepackage[toc,page,header]{appendix}
\usepackage{minitoc}
\usepackage{bbding}  
\usepackage{pifont}  
\usepackage{float}
\usepackage{multirow}
\usepackage{multicol}
\usepackage{xspace}
\usepackage{booktabs}
\usepackage{colortbl}
\usepackage{subcaption}
\usepackage{makecell}

\definecolor{ourcolor}{HTML}{99e0eb}
\definecolor{ourblue}{HTML}{27a2c3}
\definecolor{tablecolor}{HTML}{ccf2f5}  
\definecolor{tablecolor2}{HTML}{ffcdb4}
\definecolor{citecolor}{HTML}{fe7b5b}
\definecolor{grey}{rgb}{0.9, 0.9, 0.9}
\definecolor{gred}{rgb}{0.859,0.267,0.216}
\definecolor{ggreen}{rgb}{0.059,0.616,0.345}
\definecolor{deepblue}{HTML}{27a2c3}
\definecolor{titleblue}{HTML}{0685f4}
\definecolor{deepred}{HTML}{fe7b5b}

\newcommand{\dd}[2]{\ensuremath{#1\scriptstyle{\pm#2}}}                                 
\newcommand{\ddbf}[2]{\cellcolor{tablecolor}\ensuremath{\mathbf{#1\scriptstyle{\pm#2}}}} 

\newcommand{\cc}[1]{\ensuremath{#1}}                                                    
\newcommand{\ccbf}[1]{\cellcolor{tablecolor}\textbf{#1}}                                

\usepackage[final]{corl_2025} 
\newcommand{\highlight}[1]{\textcolor{green}{#1}}

\renewcommand\highlight[1]{#1}

\title{Planning from Point Clouds over Continuous Actions for Multi-object Rearrangement}

\author{
  Kallol Saha$^{*1}$, \ Amber Li$^{*1}$, \ Ángela Rodriguez-Izquierdo$^{*2}$, \ Lifan Yu$^{1}$, \\[5px]
  \textbf{Ben Eisner$^{1}$, \ Maxim Likhachev$^{1}$, \ David Held$^{1}$} \\[5px]
  $^1${Robotics Institute, Carnegie Mellon University} \ \ \ \ \ \
  $^2${Princeton University} \\[3px]
  \small{$^*$Equal Contribution}
}

\begin{document}
\doparttoc
\faketableofcontents

\maketitle
\vspace{-30px}

\begin{abstract}
Long-horizon planning for robot manipulation is a challenging problem that requires reasoning about the effects of a sequence of actions on a physical 3D scene. 
While traditional task planning methods are shown to be effective for long-horizon manipulation, they require discretizing the continuous state and action space into symbolic descriptions of objects, object relationships, and actions. 
Instead, we propose a hybrid learning-and-planning approach that leverages learned models as domain-specific priors to guide search in high-dimensional continuous action spaces.
We introduce \textbf{SPOT}: \textbf{S}earch over \textbf{P}oint cloud \textbf{O}bject \textbf{T}ransformations, which plans by searching for a sequence of transformations from an initial scene point cloud to a goal-satisfying point cloud.
SPOT samples candidate actions from learned suggesters that operate on partially observed point clouds, eliminating the need to discretize actions or object relationships.
We evaluate SPOT on multi-object rearrangement tasks, reporting task planning success and task execution success in both simulation and real-world environments.
Our experiments show that SPOT generates successful plans and outperforms a policy-learning approach. We also perform ablations that highlight the importance of search-based planning.

\textbf{Website:}
\url{https://planning-from-point-clouds.github.io}
\end{abstract}
\vspace{-10px}

\vspace{-8px}


\begin{figure}[!h]
    \centering
    \includegraphics[width=\linewidth]{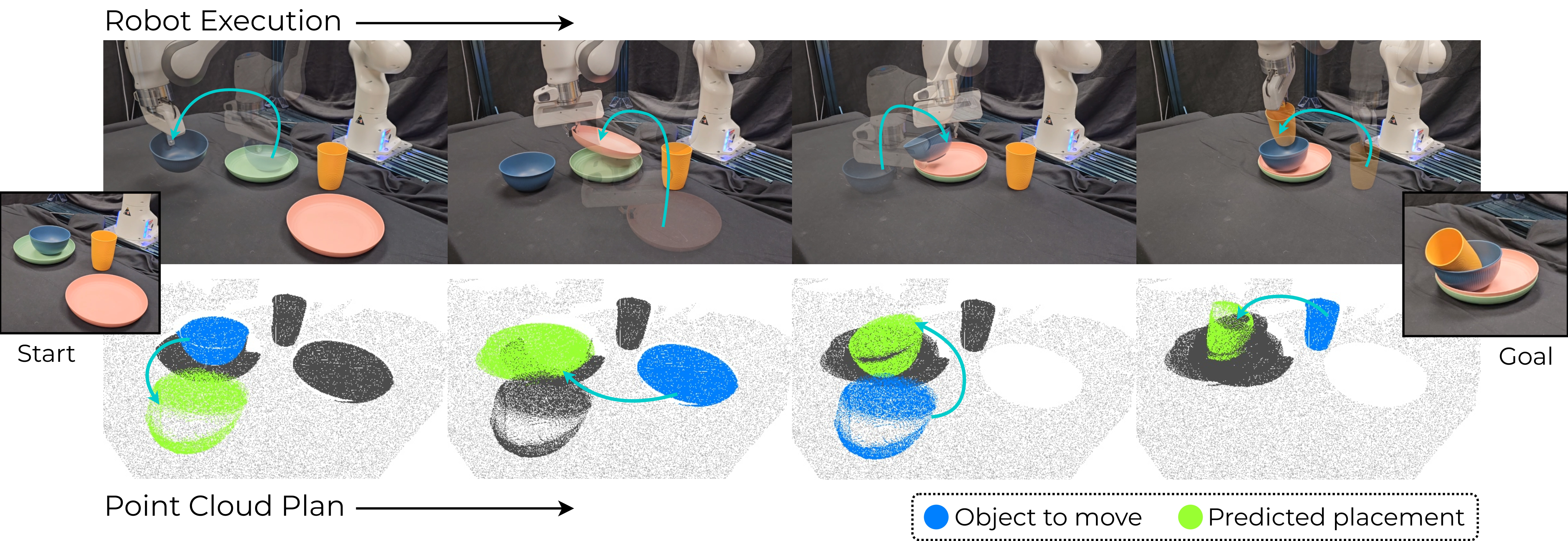}
    \caption{
    \textbf{SPOT: Search over Point cloud Object Transformations.}
    Our method solves multi-object rearrangement tasks by planning directly in point cloud space, without using any privileged ground-truth information such as the ground-truth object states. Starting from a segmented point cloud, it performs A* search over object-wise $SE(3)$ transformations until a goal configuration is found. The output plan is then executed in the real world.}
    \label{fig:teaser}
    \vspace{-5mm}
\end{figure}


\vspace{10px}
\section{Introduction}

Long-horizon robot manipulation tasks require understanding a 3D physical scene, and reasoning about the effects of a sequence of actions.
For example, in multi-object rearrangement tasks, where a robot must move a set of objects from an initial configuration into a goal-satisfying configuration (Figure~\ref{fig:teaser}), several challenges arise.
A task like table bussing -- re-stacking objects such as plates, cups, and bowls -- can be challenging because the robot may need to carefully unstack items before assembling them into the desired stacking configuration.
For example, before stacking a plate on a plate, any bowls or cups on top of it must be first removed and set aside.
Similarly, packing objects onto a shelf can be challenging when space is limited, since the exact placement of each item determines whether everything will fit.
Planning becomes essential in orchestrating such long-horizon operations.


    
Symbolic planning~\cite{kaelbling2013integrated, pasula2007learning, 5509563} is a classical family of methods for long-horizon robot manipulation tasks.
However, these approaches require a symbolic scene description, including a symbolic description of all objects, object relationships, actions, preconditions, and post-conditions.
Obtaining such definitions for a real-world scene is challenging, as these approaches assume total knowledge of the scene state and dynamics. 
Further, such knowledge is task-specific and often requires human annotations for each  task; making them difficult to apply to open-set problems.

In this paper, we propose \textbf{S}earch over \textbf{P}oint cloud \textbf{O}bject \textbf{T}ransformations (\textbf{SPOT}), a hybrid learning and search-based planning method for solving multi-object rearrangement tasks. SPOT operates on partially observed point clouds and reasons over a sequence of continuous-valued actions, without the need to discretize the action space or object relationships. 
Searching over high-dimensional and long-horizon state-action transitions is notoriously difficult; therefore, we leverage domain-specific priors learned from task demonstrations to reduce the complexity of the search problem while retaining a continuous, high-dimensional search space.  

We formulate multi-object rearrangement as a sequential partial point cloud rearrangement problem, using A* search to plan over object-level spatial transformations of the scene until a goal-satisfying configuration is found. To efficiently guide search over continuous state-action transitions, SPOT leverages two learned components: an object suggester and a placement suggester. The object suggester predicts which object(s) in the current scene can be feasibly moved, while the placement suggester, given a selected object, samples potential transformations that are executable via motion planning. Together, the suggesters guide node expansion in A* search by answering two high-level questions: \textit{which object should be moved, and where should it be placed?}
A learned model-deviation estimator (MDE)~\cite{lagrassa2022learningmodelpreconditionsplanning, McConachie_2020, power2021simpledataefficientlearningcontrolling} discourages unlikely transitions by estimating the deviation between the expected and true state. 
Together, these modules 
enable search-based planning to solve complex tasks directly from point cloud observations, eliminating the need to manually define the task ontology or rely on simulator-specific knowledge. The contributions of this paper include:\\[2px]
(1) Introducing a novel paradigm for solving rearrangement tasks without discretization, namely by planning over 3D transformations of raw 3D observations with A*. \\[2px]
(2) Learning object suggesters and 3D relative placement suggesters from video demonstrations, to aid node expansion during A* search. \\[2px]
(3) Comprehensive experiments on several real and simulated tasks, demonstrating the effectiveness of using our hybrid planning-learning approach to solve multi-object rearrangement tasks.

\section{Related Work}
\label{sec:related-work}

\subsection{Planning From Point Clouds}
Previous works make use of relational state abstractions to convert raw point cloud observations to a symbolic state for planning~\cite{huang2024points2plans, paxton2021predictingstableconfigurationssemantic, lin2023planningspatialtemporalabstractionpoint}.
Others have opted to encode the scene into a learned latent space in which to plan~\cite{huang2023erdtransformer, 8653875, huang2024latentspaceplanningmultiobject}.
In contrast to both of these approaches, SPOT does not plan in a latent space, and it is not limited to a discrete set of actions or object relationships.
\citet{simeonov2020long} also plan directly from sensory observations; however, they assume the robot is given a high-level plan skeleton consisting of 3D transformations as subgoals, whereas SPOT seeks to discover the plan through search.
\citet{structformer2022} also plan for multi-object rearrangement directly from a point cloud of the scene.
However, a key component of their approach is a language command that specifies multi-object relational constraints for a goal configuration.
SPOT is not conditioned on language, but instead uses learned knowledge about relational object placements to guide planning.

\subsection{Multi-Object Rearrangement}

Previous works have taken a variety of approaches to the problem of multi-object arrangement.
One common approach is skill sequencing, which usually leverages sampling-based optimization~\cite{AgiaMigimatsuEtAl2023, LinAgiaEtAl2023, mishra2023generative, xue2024logicskillprogrammingoptimizationbasedapproach, DBLP:journals/corr/abs-2011-04828}.
A related category of approaches has been language-guided planning~\cite{tang2025zeroshotroboticmanipulationlanguageguided,huang2024points2plans, LinAgiaEtAl2023, structformer2022, rana2023sayplangroundinglargelanguage}.
These works generally assume access to a library of skills and a library of predicates, which the language model uses to predict a set of goal predicates and generate plans that satisfy skill pre- and post-conditions.
In contrast, an underlying assumption of our method is that the robot possesses a single ``skill" -- transforming an object from an initial pose to a target pose.
Others have approached this problem by planning in a latent space using relational classifiers and a learned dynamics model~\cite{huang2023planningmultiobjectmanipulationgraph, huang2023erdtransformer}.
Conversely, SPOT does not assume a discrete set of object relationships, nor do we plan in a latent space.

\subsection{Single Object Pose Reconfiguration}

Single-object pose reconfiguration plays an important role in contact-rich manipulation tasks and is a core component of our framework.
Specifically, we leverage previous work by \citet{wang2024taxposed} that views the problem as a multi-modal relative placement task.
Other approaches seek to optimize the sequencing of manipulation primitives along with learned skill parameter samplers~\cite{paxton2021predictingstableconfigurationssemantic, simeonov2020long}.
In contrast to these works, we do not assume access to a set of skills but instead view the problem as a search over possible object configurations.
Further, unlike previous works on single-object pose reconfiguration~\cite{simeonov2023rpdiff,jiang2024hacmanpp, zhou2023hacman, fu20226droboticassemblybased, yu2024maniposecomprehensivebenchmarkposeaware}, we focus on planning for multi-object rearrangement.

\section{Problem Statement and Assumptions}
\label{sec:problem}


We tackle the problem of multi-object rearrangement, where we need to find a sequence of object pose reconfigurations to achieve a given objective.
Our observation space $S \subseteq \mathbb{R}^{n \times 3}$ is defined as a partially-observed 3D point cloud $\mathbf{o} \in S$ of the scene, segmented into a finite set of $M$ objects $\mathbf{X} =\{\mathbf{x}_1, \mathbf{x}_2, \dots, \mathbf{x}_{M} \}.$ We use an image segmentation method~\cite{sam2} combined with an object detector~\cite{dino, ren2024grounded} to assign a semantic class to each point.
We do not assume access to any ground-truth information about the scene, such as object poses.
Our action space is object-centric and continuous, where an action is defined as applying an $SE(3)$ transformation $\mathbf{T} \in \mathbb{R}^{4 \times 4}$ to an object $\mathbf{x}$.
This can be visualized by applying $\mathbf{T}$ to the point cloud of $\mathbf{x}$.
Therefore, an action $\mathbf{a}_t$ at time step $t$ is defined as a pair of an object and a transformation, $\mathbf{a}_t = \langle \mathbf{x}_t , \mathbf{T}_t \rangle$.

A task is completed when the point cloud of the scene satisfies a goal condition.
We assume access to a goal function $\mathcal{G}: S \mapsto \{0, 1\}$ that operates on a point cloud observation $\mathbf{o}\in S$, where $1$ indicates that the point cloud satisfies the goal condition.
Given an initial segmented point cloud $\mathbf{o}_1$, our aim is to compute a plan $P = (\mathbf{a}_1, \mathbf{a}_2, \dots \mathbf{a}_{T-1} )$, which when executed reaches a final point cloud $\mathbf{o}_T$ such that $\mathcal{G}(\mathbf{o}_T) = 1$, if such a plan exists.

We assume access to a limited demonstration dataset of domain-specific point cloud transitions $D = \{\mathbf{o}_i, \mathbf{a}_i, \mathbf{o}_{i+1}\}_{i=1}^{N_D}$.
This dataset is extracted directly from RGB-D videos of human demonstrations. 
It is important to note that the demonstrations are not task-specific, i.e., our planner's goal may differ from the goal in the demonstration while the objects and environment remain largely the same.
See Appendix~\ref{app:experimental_setup} for more details on the dataset.


\section{SPOT: Search over Point Cloud Object Transformations}
\label{sec:methods}

\begin{figure*}[t]
    \centering
    \includegraphics[width=0.9\textwidth]{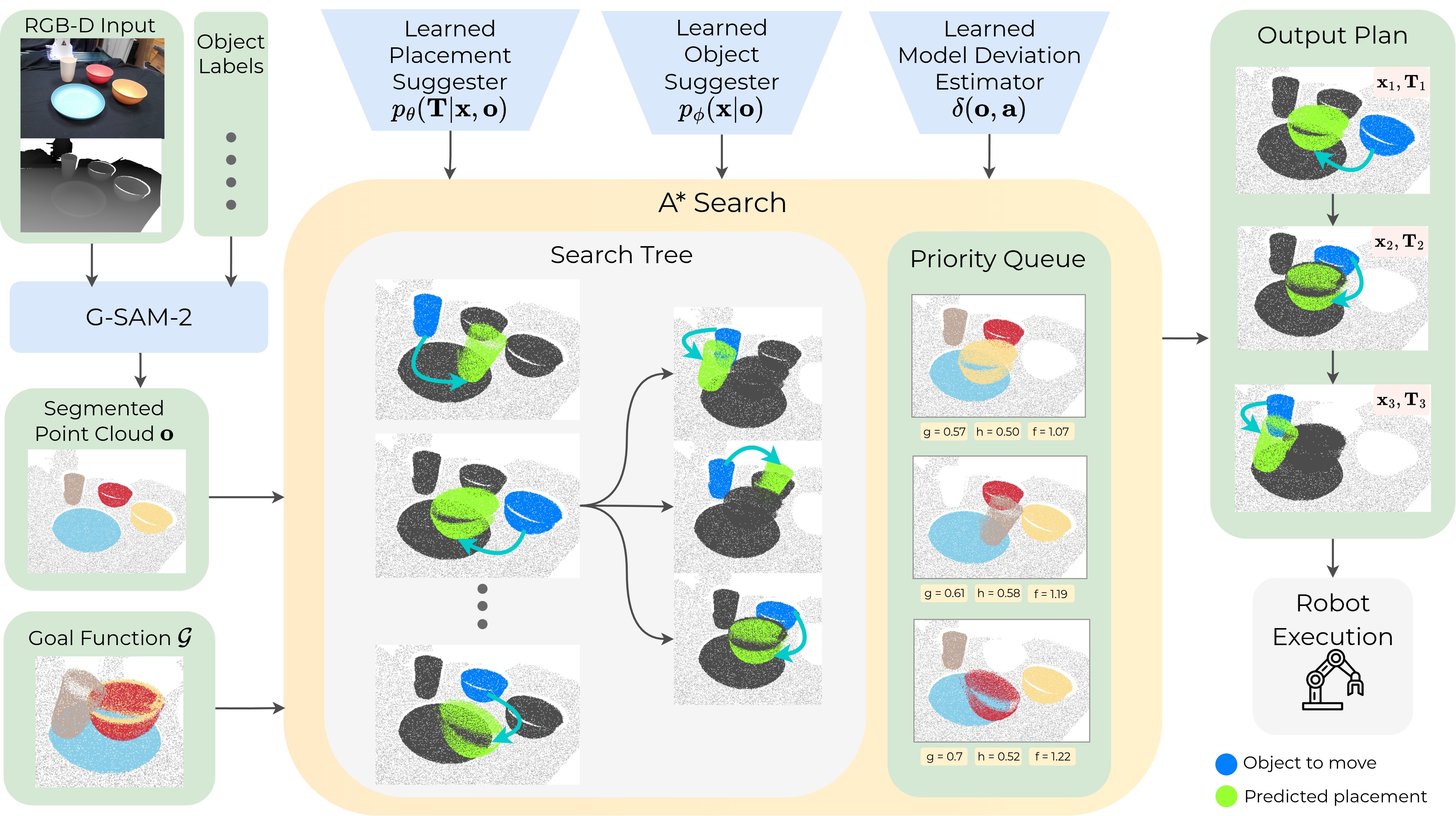}
    \caption{
    \textbf{System overview.}
    Our method takes an RGB-D image, object names, and a goal function as input. It generates a segmented point cloud and uses A* search to plan over continuous actions (object transformations) until a goal-satisfying point cloud is found. Node expansion occurs by sampling an object to move from the object suggester, then sampling a corresponding transformation from the placement suggester. The model deviation estimator biases search toward physically plausible actions. The output is a sequence of object transformations, which is given to the robot for execution.}
    \label{fig:system}
\end{figure*}

We present a novel framework for multi-object robotic manipulation tasks, integrating A* search with 3D relative placement predictions to plan over continuous state and action spaces.
As shown in Figure \ref{fig:system}, given an initial point cloud observation $\mathbf{o}_1$ and a goal function $\mathcal{G}$, we aim to compute a point cloud plan $P = ( \mathbf{a}_1, \mathbf{a}_2, \dots \mathbf{a}_{T-1} )$ using A* search over point cloud transformations (Sec.~\ref{sec:astar-search}).
Executing the plan results in a sequence of point cloud observations $O =( \mathbf{o}_2, \mathbf{o}_3, \dots, \mathbf{o}_T )$, such that $\mathcal{G}(\mathbf{o}_T) = 1$.
We formulate this using a Markovian model such that an observation $\mathbf{o}_{t+1}$ depends only on the previous observation $\mathbf{o}_{t}$ and action $\mathbf{a}_{t}$: 
\begin{equation*}
    p(O, P| \mathbf{o}_1 ) = p(\mathbf{o}_2, \dots, \mathbf{o}_T, \mathbf{a}_1, \dots, \mathbf{a}_{T-1} | \mathbf{o}_1 )
    = \prod_{t=1}^{T-1} p(\mathbf{o}_{t+1} | \mathbf{o}_{t}, \mathbf{a}_{t}) p(\mathbf{a}_{t} | \mathbf{o}_{t})
\end{equation*}
The term $p(\mathbf{o}_{t+1} | \mathbf{o}_{t}, \mathbf{a}_{t})$ represents point cloud dynamics, modeled as a transform $\mathbf{T}_t$ applied to an object $\mathbf{x}_t$ in the point cloud scene $\mathbf{o}_{t}$. The term $p(\mathbf{a}_{t} | \mathbf{o}_{t})$ can be broken down further as:
\begin{equation*}
    p(\mathbf{a}_{t} | \mathbf{o}_{t}) = p(\mathbf{x}_t, \mathbf{T}_t | \mathbf{o}_{t})
    = p(\mathbf{x}_t | \mathbf{o}_{t}) \ p(\mathbf{T}_t | \mathbf{x}_t, \mathbf{o}_{t})
\end{equation*}
where $\mathbf{x}_t$ and $\mathbf{T}_{t}$ represent the object and the transformation, respectively, that define action $\mathbf{a}_{t}$.
We learn an object suggester $p_{\phi}(\mathbf{x}_t | \mathbf{o}_{t})$ (Sec.~\ref{sec:object-suggester}) and a placement suggester $p_{\theta}(\mathbf{T}_t | \mathbf{x}_t, \mathbf{o}_t)$ (Sec.~\ref{sec:learned-placement-suggester}) from an environment-specific dataset of point cloud transitions $D$.
We sample actions from the learned object and placement suggesters, thereby expanding the search graph from the initial point cloud observation $\mathbf{o}_1$ through a finite subset of states.
Search terminates when one or more goal nodes are found, or when a maximum node expansion limit is exceeded.

After finding a plan $P$ to reach a point cloud $\mathbf{o}_T$ that satisfies the goal function, a robot manipulator sequentially executes each action $\mathbf{a} = \langle \mathbf{x}, \mathbf{T} \rangle$ of the resulting plan.
We detect grasp poses for each object $\mathbf{x}$ using Contact-GraspNet~\cite{sundermeyer2021contact} in the real world or using grasp heuristics in simulation. 
Once grasped, we move the object by transformation $\mathbf{T}$ using a robot controller.
Our method can operate from RGB-D cameras along with the names of the objects in the scene, without access to any other ground-truth information describing the scene.

Oftentimes, the executed plan differs from the output plan due to environmental dynamics. To address this, we train a model deviation estimator~\cite{lagrassa2022learningmodelpreconditionsplanning, McConachie_2020, power2021simpledataefficientlearningcontrolling} (Sec.~\ref{sec:mde}) that predicts the deviation between a planned action in the search tree and its actual execution. We use this estimate to guide the A* search towards actions with lower deviation.

\subsection{Learned object suggester}
\label{sec:object-suggester}

\begin{figure}[t]
\centering
    \begin{minipage}{\linewidth}
        \centering
        \includegraphics[width=\linewidth]{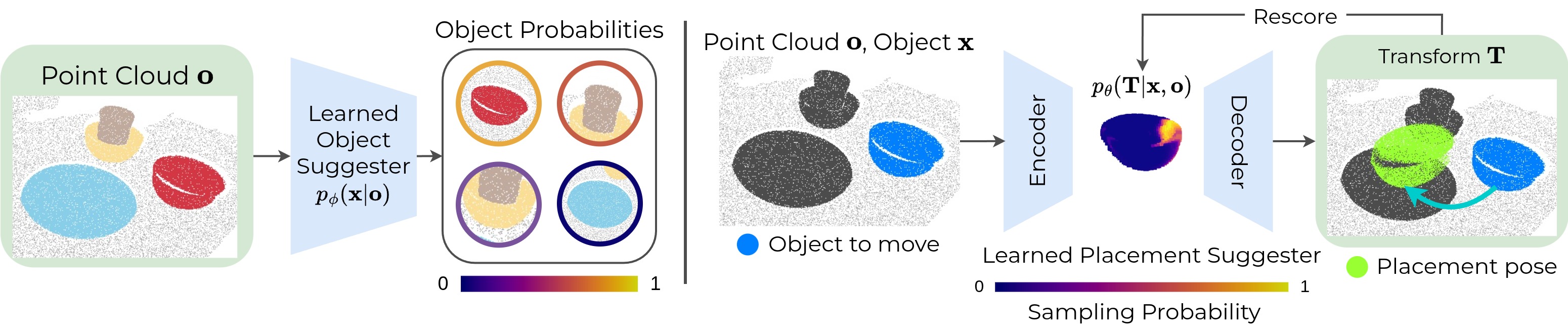}
        \captionof{figure}{\textbf{Learned object and placement suggesters}. Left: The object suggester predicts a probability distribution over which objects in the scene can be feasibly moved, given a point cloud observation of the scene. Right: Given an object and the scene point cloud, the placement suggester samples candidate transformations indicating where the object could be moved next.}
        \label{fig:obj_suggester}
    \end{minipage}
    \vspace{-13px}
\end{figure}

The object suggester guides the search by learning a distribution $p_\phi(\mathbf{x} | \mathbf{o})$ over which object $\mathbf{x}$ to move next.
This model is trained using an environment-specific offline dataset $D$ consisting of point cloud transitions.
As mentioned previously, the demonstrations in this offline dataset are not task-specific, i.e., our planner’s goal may differ from the goal in the demonstration.
Thus, the object suggester learns which objects in the current scene can feasibly be moved.

At inference time, the model outputs a score $\tilde{p}_{\mathbf{x}_i}$ for each object $\mathbf{x}_i$ in the scene: 
$\{\tilde{p}_{\mathbf{x}_i} \, | \, \mathbf{x}_i \in \mathbf{X} \}$.
We normalize these scores into the probability distribution $p_\phi(\mathbf{x})$ that predicts the likelihood that the robot should move object $\mathbf{x}$ given the current observation (Figure~\ref{fig:obj_suggester}).
Implementation details can be found in Appendix~\ref{app:object_suggester}.

\subsection{Learned placement suggester}
\label{sec:learned-placement-suggester}

Our method expands a node $n$ in the search tree into multiple child nodes leveraging a placement suggester $p_{\theta}(\mathbf{T} | \mathbf{x}, \mathbf{o})$ trained from demonstration data.
For example, in a table bussing environment, given that object $\mathbf{x}$ is a cup, the placement suggester may suggest transformations that place the cup on top of a plate or inside of a bowl, resulting in two different possible child nodes of an expanded node.

To train the suggester, we also use dataset $D$.
Recall that for each point cloud transition in $D$, there is some action $\mathbf{a}_i = \langle \mathbf{x}_i, \mathbf{T}_i \rangle$ that is applied to a point cloud observation $\mathbf{o}_i$.
The placement suggester learns the distribution $p_{\theta}(\mathbf{T} | \mathbf{x}, \mathbf{o})$ of possible transformations $\mathbf{T}$ of an object $\mathbf{x}$ throughout the task (Figure~\ref{fig:obj_suggester}).
At inference time, the placement suggester can sample multiple task-relevant transformations for each object in a given scene, generating a batch of possible child nodes.
In practice, our placement suggester is implemented as TAXPose-D~\cite{wang2024taxposed}, a method for relative placement tasks that accounts for multi-modality over task solutions.
For more details on TAXPose-D, see Appendix~\ref{app:placement suggesters}.

\subsection{Model deviation estimator}
\label{sec:mde}
Our placement suggester $p_{\theta}(\mathbf{T} | \mathbf{x}, \mathbf{o})$
assumes that for a given object $\mathbf{x}$, there is a transformation $\mathbf{T}$ that moves only that object and results in the next observation.
However, this may not match the actual dynamics of the environment.
For instance, the placement suggester may propose moving a plate that has a cup on it, but actually executing this action would result in an observed next state that differs from the next state the suggester expects, since the cup would likely fall off the plate when the plate is moved.
To address this issue, we learn a model deviation estimator (MDE)~\cite{lagrassa2022learningmodelpreconditionsplanning, McConachie_2020, power2021simpledataefficientlearningcontrolling} and integrate it into the A* search such that it favors expanding child nodes whose transitions can be accurately predicted by the placement suggester.
The MDE $\delta (\mathbf{o}, \mathbf{a})$ takes in a point cloud observation $\mathbf{o}$ and the suggested action $\mathbf{a}$, and outputs a real number as the estimated deviation.
More implementation details can be found in Appendix~\ref{app:mde}.

\subsection{A* search over scene configurations}
\label{sec:astar-search}

Given an initial point cloud observation $\mathbf{o}_1$, finding the shortest plan that reaches some final point cloud $\mathbf{o}_T$ such that $\mathcal{G}(\mathbf{o}_T)=1$ can be formulated as a search problem in the space of all possible object rearrangements. 

We tackle this planning problem using A* search.
We define a node $n$ in the search tree as a combination of a point cloud $\mathbf{o}_n$, an action $\mathbf{a}_n = \langle \mathbf{x}_n, \mathbf{T}_n \rangle $, and its parent node $n_p$, written as $n = [\mathbf{o}_n, \mathbf{x}_n, \mathbf{T}_n, n_p]$.
A node $n$ is derived from its parent $n_p$ by applying the action $\mathbf{a}_n$ to the parent's point cloud $\mathbf{o}_{n_{p}}$.
We assume a directed search graph where every node except the initial configuration has a single parent.

Starting from the initial point cloud $\mathbf{o}_1$ contained in the initial node $n_1$, the A* search algorithm maintains a tree of paths, iteratively expanding new nodes by sampling transformations until a goal is found.
A child node $n' = [\mathbf{o}_{n'}, \mathbf{x}_{n'}, \mathbf{T}_{n'}, n]$ is generated from the current node $n$ by sampling an object $\mathbf{x}_{n'}$ and a transformation $\mathbf{T}_{n'}$:
$$\mathbf{a}_{n'} = \langle \mathbf{x}_{n'} , \mathbf{T}_{n'} \rangle \sim \langle \ p_{\phi}(\mathbf{x}_{n'} | \mathbf{o}_n) \ , p_{\theta}(\mathbf{T}_{n'} | \mathbf{x}_{n'}, \mathbf{o}_n) \ \rangle$$
We propose $k$ potential transformations for every object $\mathbf{x}$ in the set of all objects $\mathbf{X}$ in the scene, where $k$ is a hyperparameter.
We use the learned placement suggester $p_{\theta}(\mathbf{T} | \mathbf{x}, \mathbf{o})$ (Sec.~\ref{sec:learned-placement-suggester}) to propose a distribution of placements of where to move object $\mathbf{x}$. 

\subsubsection{Value function}
\label{sec:cost-function}
\label{sec:heuristic}
Proper guidance toward the goal becomes crucial in high-dimensional domains such as ours, which consists of all possible configurations of the set of objects in the scene.
The A* search algorithm prioritizes expanding nodes that are more likely to be closer to the goal, according to a value function $f(n) = g(n) + h(n),$ where $g(n)$ is the cost function, and $h(n)$ is the heuristic function.

The cost function $g(n)$ of a node $n$ is defined as a weighted sum: 
$$g(n) = C_a(n) + w_cC_c(n) + w_dC_d(n) + w_pC_p(n)$$ 
An \textbf{action cost} $C_a(n)$ approximates the step cost needed to reach the current node from the root node through a path.
A \textbf{collision cost} $C_c(n)$ discourages suggested placements that may result in collision, e.g. placing a cup improperly such that it collides with a bowl.
A \textbf{deviation cost} $C_d(n)$ guides search away from transitions unlikely to match the actual dynamics of the environment when executed by the manipulator.
This is predicted by a learned model deviation estimator (Sec.~\ref{sec:mde}).
A \textbf{probability cost} $C_p(n)$ favors transitions that are more likely to be seen in the demonstration dataset $D$.
This is derived from the outputs of the learned placement suggester and the object suggester (Sec.~\ref{sec:object-suggester}, \ref{sec:learned-placement-suggester}).

The heuristic function $h(n)$ is a task-specific function over the segmented point cloud $\mathbf{o}_n$ of a node $n$. In practice, we let A* search for multiple goals, rather than terminating at the first goal found.
More details on our search method, cost, and heuristic functions can be found in Appendix~\ref{app:nodes}.

\section{Results}
\label{sec:results}

\begin{figure}[t]
\centering
    \begin{minipage}{\linewidth}
        \centering
        \includegraphics[width=\linewidth]{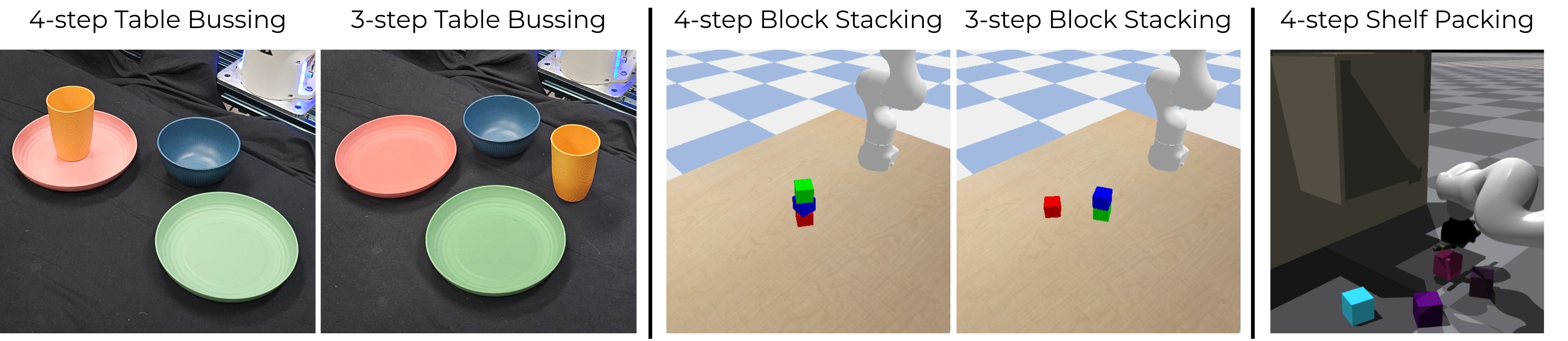}
        \captionof{figure}{\textbf{Initial configurations.} We use initial configurations of varying complexity for each task.
    In a 4-step table bussing task, the robot first has to remove the cup from the plate before stacking all objects. For block stacking, where the goal is to arrange blocks in the order red, green, blue from top to bottom, a 4-step task involves unstacking all blocks before restacking them in the correct order. The constrained packing task requires a robot to grasp and place four blocks inside a shelf.}
        \label{fig:configurations}
    \end{minipage}
    \vspace{-13px}
\end{figure}

We show experiments demonstrating that we can perform planning from point cloud observations using only learned priors from domain-specific datasets.
In our experimental evaluation, we aim to answer the following questions:
(1) Can our method produce goal-satisfying plans after training only on point cloud demonstrations?
(2) Do these plans achieve task success reliably when executed on a robot?

To answer these questions, we evaluate SPOT on a variety of tasks in simulation and real-world setups with initial configurations shown in Figure~\ref{fig:configurations}.

\textbf{Block Stacking (Simulation).} The robot is tasked with picking and placing cubes in the right order to a desired goal stack configuration, while aligning them precisely to prevent toppling.
The cubes may be initialized in random stacked or unstacked configurations, requiring the robot to plan both the unstacking and restacking order. The initial positions and orientations of the blocks are randomized.

\textbf{Constrained Packing (Simulation).}
To complete this task, the robot must place items into a spatially constrained environment (a cupboard) by carefully reasoning about the objects' planned placement positions.
We define success as when all of the objects are arranged inside the cupboard and are not stacked on top of each other.
Please refer to prior work by~\citet{huang2024points2plans} for more details.

\textbf{Table Bussing (Real World).} This task involves clearing a table of dinnerware by grasping and placing the objects (plates, bowls, cups) such that they are all arranged on top of one plate.
The initial positions are randomized, i.e. objects may be stacked together, requiring the robot to unstack objects before rearranging them into the goal configuration.

Further details on initial scene configurations, task complexity, and datasets for these tasks are provided in Appendix~\ref{app:experimental_setup}.

\subsection{Evaluation}
We evaluate SPOT based on the \textbf{task planning success rate}.
A task is considered complete when all objects in the current scene are arranged in the desired goal configuration according to our goal function~(Sec.~\ref{sec:problem}).
To show the efficacy of our plans, we also evaluate SPOT based on the \textbf{task execution success rate}.
For both simulation and real-world tasks, the predicted actions are executed in the corresponding environment by a Franka arm using pick-and-place primitives; in the real world, grasp poses are generated using Contact-GraspNet~\cite{sundermeyer2021contact}.

\subsection{Baselines and Ablations}
We compare our method to the following baseline and ablations:

\textbf{Points2Plans.}
Similar to our approach, prior work by~\citet{huang2024points2plans} along with ~\cite{huang2023erdtransformer, paxton2021predictingstableconfigurationssemantic} plans to solve long-horizon manipulation tasks from partially-observed point clouds. 
In contrast to SPOT, these methods make use of relational state abstractions to convert the point cloud observations to a symbolic representation of the scene for planning.
We compare against these approaches by running SPOT in their constrained packing environment.

\textbf{3D Diffusion Policy}~\cite{Ze2024DP3} is an end-to-end imitation learning policy.
Our 3D Diffusion Policy baseline is trained on a dataset of 23 expert task-specific demonstrations of block stacking provided by a human expert.
All of the demonstrations lead to a single consistent goal configuration since 3D Diffusion Policy is not goal-conditioned.
More details on the dataset, training, and evaluation can be found in Appendix~\ref{app:dp3_baseline}.

\textbf{Beam Search.} Using a beam width of 1, at each step the policy chooses the child node with the best heuristic value from the pool of all candidates that SPOT would generate during A* search.
Thus, this policy ablates the search component while receiving task-specific guidance via the heuristic.

\textbf{Random Rollouts.}
We replace A* search with a series of random rollouts, to ablate whether the improvement is merely a result of the number of nodes expanded.
This method chooses a child node uniformly at random from a set of possible child nodes.
We continue searching until the rollout exceeds the node expansion budget that we use for SPOT.
See Appendix~\ref{app:random_rollouts} for more details. 

\textbf{No Object Suggester.} 
Recall that the probability cost in the A* cost function is calculated using $p_\phi(\mathbf{x} | \mathbf{o})$, the probability of selecting object $\mathbf{x}$ to move from a point cloud $\mathbf{o}$, as predicted by the learned object suggester (Sec.~\ref{sec:object-suggester}).
We ablate the object suggester by replacing this distribution with a uniform probability distribution over all of the objects in the scene.
\begin{table}[t]
\begin{minipage}{\textwidth}
\centering
\resizebox{0.8\textwidth}{!}{%
\begin{tabular}{l|cc|cc}
\toprule
& \multicolumn{2}{c|}{\textbf{Planning Success (\%)} ($\uparrow$)} & \multicolumn{2}{c}{\textbf{Execution Success (\%)} ($\uparrow$)} \\[5px]

Algorithm \textbackslash{} Task & Block Stacking & Table Bussing & Block Stacking & Table Bussing \\
\midrule
Random Rollouts &  \dd{26}{7} & \cc{57} & \dd{21}{3} & \cc{14} \\
No Object Suggester &  \dd{86}{7} & \cc{93} & \dd{60}{7} & \cc{64} \\
\cellcolor{tablecolor}\textbf{SPOT (Ours)} &  \ddbf{88}{2} & \ccbf{100} & \ddbf{63}{4} & \ccbf{86} \\
\bottomrule
\end{tabular}}
\vspace{5px}
\caption{\textbf{Planning \& execution success rates}. We compare SPOT to random rollouts and ablate the object suggester. For block stacking, results are averaged over 5 seeds and show a 95\% confidence interval.}
\label{tab:results_main}
\end{minipage}
\end{table}

\begin{figure}[t]
\centering
    \vspace{-20px}
    \begin{minipage}{0.45\linewidth}
        \centering
        \small
        \captionsetup{type=table}
        \resizebox{\linewidth}{!}{%
        \begin{tabular}{l|ccc}
        \toprule
        & \multicolumn{3}{c}{\textbf{Execution Success (\%)} ($\uparrow$)}\\[3pt]
        Alg$\backslash$ Complexity & 2 steps & 3 steps & 4 steps \\
        \midrule
        3D Diffusion Policy & \cc{13} & \cc{0} & \cc{0} \\
        Beam Search         & \cc{43} & \cc{7} & \cc{0} \\
        \cellcolor{tablecolor}\textbf{SPOT (Ours)} & \ccbf{78} & \ccbf{58} & \ccbf{30} \\
        \bottomrule
        \end{tabular}}
        \captionof{table}{\textbf{Execution success rates on block stacking tasks}. Complexity is measured by the length of a ground-truth optimal plan to reach the goal. Results are averaged over
        4 seeds for 3D Diffusion Policy and 5 seeds for all other methods.}
        \label{tab:results_baseline}
    \end{minipage}
    \hfill
    \begin{minipage}{0.53\linewidth}
        \centering
        \includegraphics[width=\linewidth]{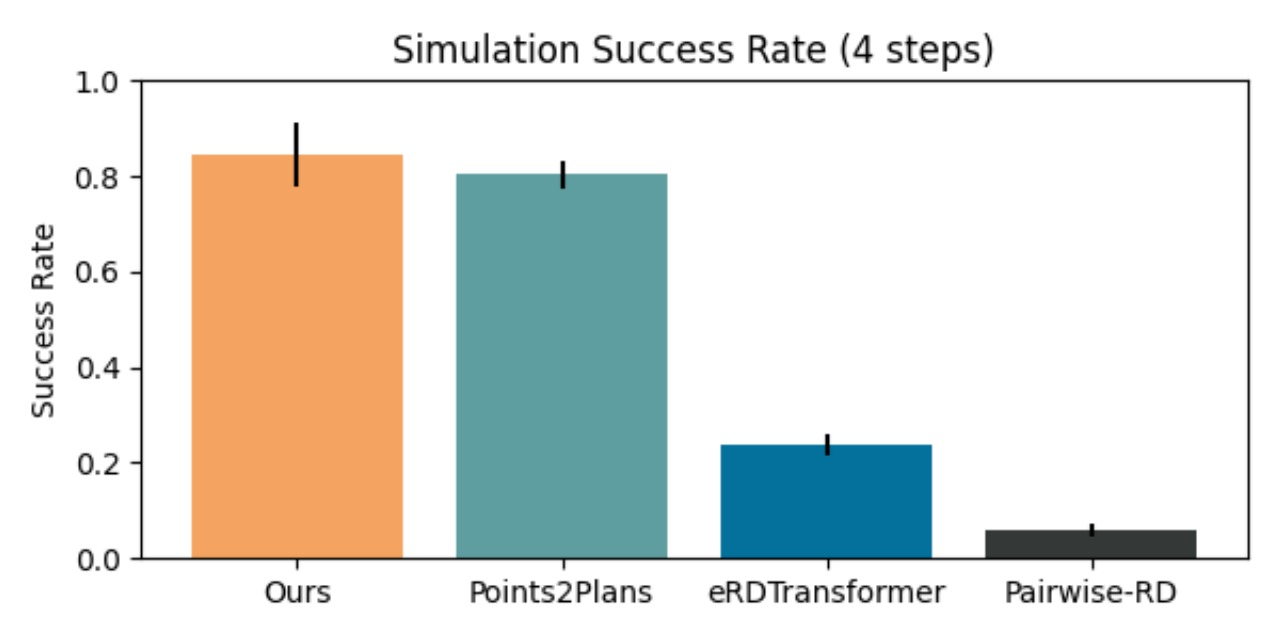}
        \vspace{-15px}
        \captionof{figure}{\textbf{Task execution success on constrained packing}. Error
bars show standard deviations across 500 trials.}
        \label{fig:points2plans_comparison}
    \end{minipage}
    \vspace{-13px}
\end{figure}

\subsection{Results \& Discussion}
\subsubsection{Simulation Results}
On block stacking, we demonstrate that SPOT produces goal-satisfying plans, many of which achieve task success when executed on a robot (Table~\ref{tab:results_main}).

\textbf{Effect of search.}
To demonstrate the benefit of search, we compare SPOT with the Beam Search ablation.
Since this method myopically chooses a single child node at each step, it cannot explore diverse actions that may lead closer to a goal configuration.
This results in a very low task planning success rate compared to SPOT (Table~\ref{tab:results_baseline}).

\textbf{Effect of guidance toward the goal.}
To demonstrate that SPOT does not achieve planning success merely by expanding a large number of possible nodes, we compare it with the Random Rollouts ablation.
SPOT is guided by A* search and its cost and heuristic functions (Sec.~\ref{sec:astar-search}), whereas Random Rollouts goes without such guidance, so our method greatly outperforms it in terms of task planning success rate (Table~\ref{tab:results_main}).

\textbf{Effect of object suggester.}
SPOT outperforms the ablation without the object suggester (Table~\ref{tab:results_main}), thus illustrating the benefits of using this component to guide our search.

\textbf{Constrained Packing Task.}
SPOT slightly outperforms Points2Plans~\cite{huang2024points2plans} on the constrained packing task, achieving 100\% task planning success and 84\% task execution success on an evaluation set consisting of 5 different initial configurations run with 100 different random seeds each (Figure~\ref{fig:points2plans_comparison}).
\highlight{
We outperform eRDTransformer~\cite{huang2023erdtransformer} and Pairwise-RD~\cite{paxton2021predictingstableconfigurationssemantic} which use an absolute dynamics model and only capture pairwise object interactions, respectively.
}
See Appendix~\ref{app:experimental_setup} for more details.

\textbf{Comparison to 3D Diffusion Policy.}
We evaluate SPOT and 3D Diffusion Policy on five tasks of varying complexity, where \textit{task complexity} is defined by the number of steps in a ground-truth optimal plan to reach a goal state (Table~\ref{tab:results_baseline}).
We see that although 3D Diffusion Policy achieves some success on 2-step tasks, it is unable to complete any 3 or 4-step tasks, while SPOT has a much higher success rate.
%

\subsubsection{Real-world Results}
We show that our method is not specific to simulation but also works in the real world by performing evaluations in the table bussing environment using SPOT trained on data collected in the real world.
The results are found in Table \ref{tab:results_main}.
For further visualization of our results, see Figure~\ref{fig:teaser} and the \href{https://planning-from-point-clouds.github.io/}{project website}.

\subsubsection{Search Metrics}
\begin{figure}[t]
\centering
    \begin{minipage}{0.37\linewidth}
        \centering
        \small
        \captionsetup{type=table}
        \resizebox{\linewidth}{!}{%
        \begin{tabular}{l|ccc}
        \toprule
        \textbf{Steps} & \makecell{\textbf{Search} \\ \textbf{Time (s)}} & \makecell{\textbf{Generated} \\ \textbf{Nodes}} & \makecell{\textbf{Expanded} \\ \textbf{Nodes}} \\
        \midrule
        1 step & \cc{1} & \cc{33} & \cc{2} \\
        2 step & \cc{11} & \cc{391} & \cc{20} \\
        3 step & \cc{32} & \cc{1123} & \cc{57} \\
        4 step & \cc{49} & \cc{1711} & \cc{86} \\
        \bottomrule
        \end{tabular}}
        \captionof{table}{\textbf{Search Metrics}. We report the computational and time costs of our search method as a function of task complexity in the block stacking environment. Figures are averages. Search time is in seconds.}
        \label{tab:metrics}
    \end{minipage}
    \hfill
    \begin{minipage}{0.61\linewidth}
        \centering
        \includegraphics[width=\linewidth]{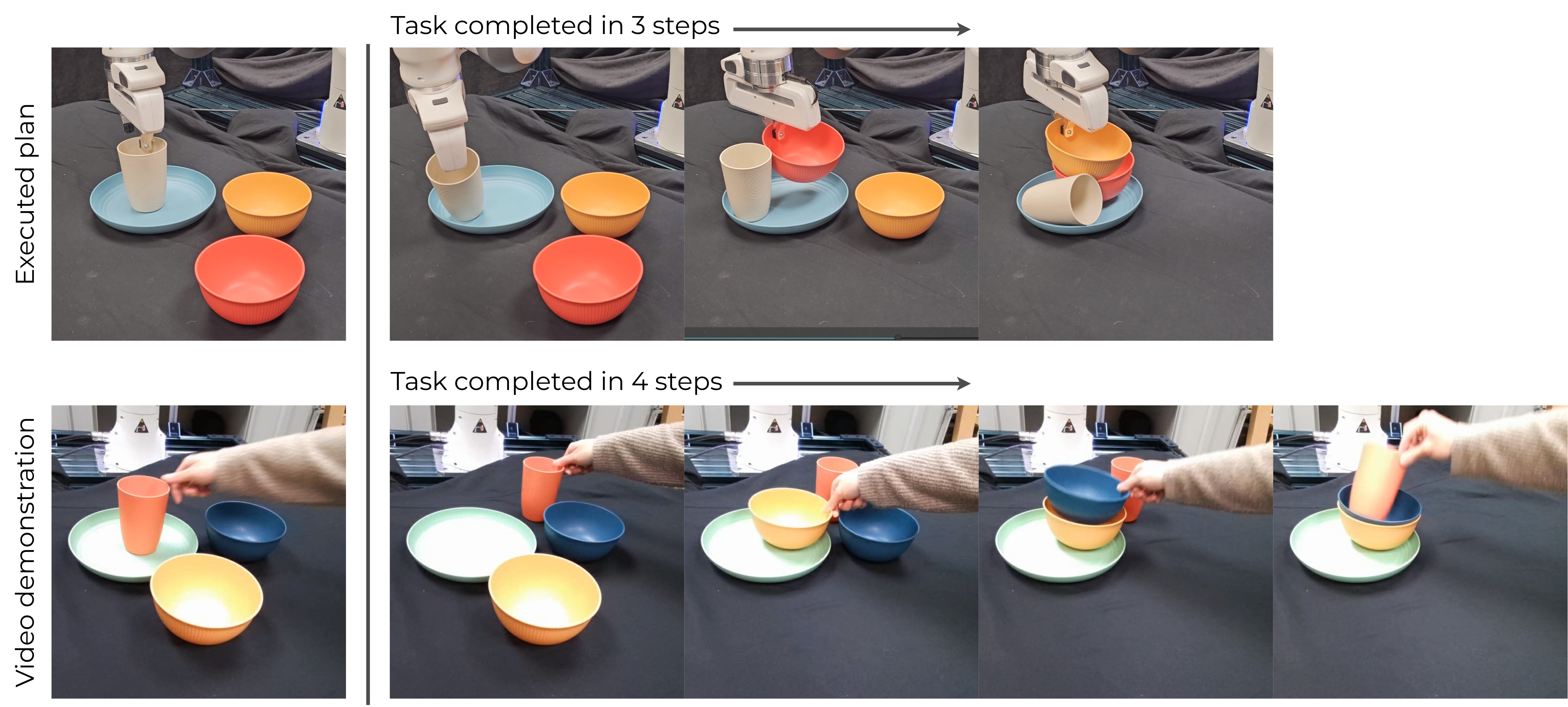}
        \vspace{-15px}
        \captionof{figure}{\textbf{Efficient Path-finding}. SPOT finds a more efficient plan than the video demonstration by first moving the cup beside the plate, creating space to stack the bowls directly. In contrast, the demonstration spends an extra step to remove the cup from the plate.}
        \label{fig:emergent}
    \end{minipage}
    \vspace{-13px}
\end{figure}

We compute the following quantitative metrics for our method:
\textbf{Total search time}: time (in seconds) that our algorithm takes to output a goal configuration.
\textbf{Generated nodes}: total number of nodes proposed during the search procedure.
\textbf{Expanded nodes}: out of the generated nodes, the number of nodes that were expanded.
Table~\ref{tab:metrics} shows the average values of these metrics for plans output by SPOT in the block stacking environment.

\subsubsection{Qualitative Analysis: Efficient Path-finding}
SPOT can often complete the task more efficiently than the human demonstrations, 
since our learned object and placement suggesters capture a goal-agnostic distribution of relevant objects and placements.
By sampling from this distribution, our approach explores different possible solutions, allowing it to find shorter and more effective paths to the goal.
Figure~\ref{fig:emergent} 
shows an example of this behavior on the table bussing task. See Appendix~\ref{app:search_graphs} for more visualizations.


\section{Conclusion} 
\label{sec:conclusion}

In this work, we presented SPOT, a hybrid learning and planning method for solving multi-object rearrangement tasks that is trained only on point cloud demonstrations.
Given a partially-observed point cloud observation of an initial scene, our method uses A* search to plan in continuous state and action spaces, without discretizing the set of actions or object relationships.
Learned, domain-specific priors from an \textit{object suggester} and a \textit{placement suggester} guide the search by proposing diverse, task-relevant object transformations from raw 3D point clouds.
A model deviation estimator guides the search towards transitions that are more likely to match planned actions.
Through experiments, we demonstrated the effectiveness of SPOT in terms of task planning and task execution success rates.
Comparisons with ablations show that guided search-based planning is critical to our method's efficacy.
\section{Limitations \& Future Work}
\label{sec:limitations}

Despite demonstrated planning success, there exist avenues for future work.
One possibility is to apply techniques geared toward searching over continuous state and action spaces, such as  Partial Expansion A*~\cite{Felner_Goldenberg_Sharon_Stern_Beja_Sturtevant_Schaeffer_Holte_2021} and Monte Carlo Tree Search. These may help SPOT find point cloud plans more efficiently.
We could also implement replanning so that the robot can recover from unexpected environment dynamics or execution failures.
The learned placement suggester could also be improved, such as by incorporating newer methods for relative placement~\cite{cai2024tax3d}.

SPOT may fail due to either planning or execution failures:
\paragraph{Planning failures.}
Planning may fail, in which case no plan is found.
Reasons for planning failures include:
\begin{itemize}
    \item The learned placement suggester may give poor suggestions, leading to a failure to find a path to the goal.
    
    \item The model deviation estimator may give poor predictions; for example, the model may erroneously output a large predicted deviation for a ``good" action, thereby misleading A* search away from states that would satisfy the goal function.

    \item The object suggester may give poor predictions of which object(s) should be moved. If the model erroneously predicts that an object should not be moved when it should, this may misguide A* search.
\end{itemize}

\paragraph{Execution failures.}
Execution may fail due to a few reasons:
\begin{itemize}
    \item The learned placement suggester may give imprecise suggestions that cannot be executed by the robot.
    
    \item Point cloud occlusions make estimating a collision cost challenging.
    The collision cost in our method (Sec.~\ref{sec:cost-function}) is meant to guide A* search away from nodes corresponding to transitions that would see some sort of object-object collision.
    However, the point clouds are partially observed, e.g., when an initial state in table bussing has the cup nested inside of a bowl.
    This occlusion makes it difficult to reliably detect collisions, which is why we opted for a collision cost, as opposed to pruning nodes during search using a threshold value.

    \item The model deviation estimator may also cause execution failures. If the model erroneously outputs a small predicted deviation for some ``bad" action, the planner may output a plan containing the bad action. Such a plan may, e.g., instruct the robot to move a plate that has a cup on top of it. During execution, the next state will deviate from that predicted by the placement suggester, leading to execution failure.

\end{itemize}


\clearpage
\acknowledgments{
\highlight{
This material is based upon work supported by ONR MURI N00014-24-1-2748 and by the Toyota Research Institute.
We are very grateful to Jacinto Suñer and Yufei Wang for their invaluable assistance with the real-world robot setup and experiments.
Many thanks to Eric Cai, Tal Daniel, Sriram Krishna, Divyam Goel, and other members of R-PAD lab and SBPL for constructive feedback on paper drafts.
We also thank Yixuan Huang for answering our questions about Points2Plans.
Finally, we thank the anonymous reviewers for their feedback.
}
}

\subsubsection*{Contributions}
\begin{itemize}
    \item \textbf{KS}: Trained and evaluated placement suggesters. Contributed to object suggester training and evaluation. Set up real-world and simulation environments for data collection and benchmarking. Designed cost function components for planning. Ran real-world experiments, ran the shelf packing benchmark, and contributed to other simulation benchmarks. Contributed to paper writing and made figures.
    \item \textbf{AL}: Trained and evaluated object suggesters and model deviation estimators. Contributed to simulation environment setup and data collection in simulation and real-world environments. Ran benchmark experiments in simulation. Contributed to paper writing.
    \item \textbf{AR}: Executed an initial version of the project with experiments using placement suggesters and A* search.
    \item \textbf{LY}: Trained and evaluated the DP3 baseline. Contributed to real-world experiments.
    \item \textbf{BE}: Mentored the project. Initially conceived of the project idea, and helped in brainstorming the method, experiments and ablations. Contributed to paper writing.
    \item \textbf{ML}: Advised the project, shaped the research direction and method, suggested experiments.
    \item \textbf{DH}: Advised the project, shaped the research direction and method, suggested experiments, provided computational and personnel resources.
\end{itemize}

\bibliography{references}  

\newpage
\appendix
\addcontentsline{toc}{section}{Appendix} 
\part{Appendix}
\parttoc 

\section{Object Suggester}
\label{app:object_suggester}
In our method, the learned object suggester predicts which object(s) in the current scene can be feasibly moved (Sec.~\ref{sec:object-suggester}).

\subsection{Dataset}
The object suggester is trained in a supervised fashion from a transition dataset whose demonstrations are \textit{not} task-specific, so as a result, the model learns which objects in the current scene can be moved feasibly.
Each piece of transition data in the dataset $D$ consists of a point cloud observation $\mathbf{o},$ an action $\langle \mathbf{x}, \mathbf{T} \rangle$ taken on $\mathbf{o}$, and the resulting next scene point cloud $\mathbf{o}'.$
For table bussing, the dataset consists of 154 transitions, and for block stacking, there are 78.

\subsection{Implementation Details}
As shown in Figure~\ref{fig:obj_suggester}, the object suggester receives a ``query mask" as input, which defines the object $\mathbf{x}$ being input to the function $p_\phi(\mathbf{x} | \mathbf{o})$.
To create this input query, we construct a binary mask over all points in the scene point cloud $\mathbf{o},$ where points belonging to object $\mathbf{x}$ receive a mask value of $1$, while points that do not belong to object $\mathbf{x}$ receive a mask value of $0$.
This input is a query mask that  essentially asks the object suggester, ``Should this object be moved next?"
The corresponding ground-truth label for this query is then $1$ since $\mathbf{x}$ is the object being moved in the demonstration.
We also construct query masks for the other objects in the scene, and those examples are labeled $0.$

At inference time, we iterate over the set of movable objects $\mathbf{X}$ in the scene, and for each object $\mathbf{x}_i$, we construct a corresponding binary mask with which to query the object suggester.
For each object $\mathbf{x}_i$, the object suggester returns a score $\tilde{p}_{\mathbf{x}_i}$.  These queries form a collection of real number outputs, $\{\tilde{p}_{\mathbf{x}_i} \, | \, \mathbf{x}_i \in \mathbf{X} \}$.
We normalize these outputs into the probability distribution $p_\phi(\mathbf{x})$ that predicts the likelihood that the robot should move object $\mathbf{x}$ given the current observation.

\subsection{Training Details}
The object suggester takes as input a point cloud observation $\mathbf{o}$ of the current scene and a binary query mask for the object $\mathbf{x}$ we want to query about, i.e., \textit{should the robot move object $\mathbf{x}$ given the current point cloud observation $\mathbf{o}$?}
The model outputs a real number. 
We use a PointNet++ network~\cite{qi2017pointnetplusplus} for the architecture.
The object suggester in each environment is trained for 2500 steps.
We use a batch size of 16 and a learning rate of 1e-3.

\section{Learned Placement Suggester}
\label{app:placement suggesters}
Given an object to move, our method samples multiple task-relevant transformations from a learned placement suggester (Sec.~\ref{sec:learned-placement-suggester}).

\subsection{TAXPose-D}
Our learned placement suggester is implemented as TAXPose-D \cite{wang2024taxposed}.
To provide some background, TAXPose-D relies on a Conditional Variational Autoencoder (cVAE) that maps the input point cloud $\mathbf{o}$ into a latent distribution $\tilde{p}(z|\mathbf{x}, \mathbf{o})$ over the cloud, conditioned on the object $\mathbf{x}$ to be moved.
This distribution can be normalized using softmax to a probability distribution $p(z|\mathbf{x}, \mathbf{o})$, from which we sample a latent encoding $z$ that is decoded into a transformation that achieves the relative placement.

TAXPose-D represents the latent space as a discrete multinomial distribution over the set of points, which is beneficial for handling discrete multi-modal placements.
For example, to place a cup inside a bowl as opposed to on top of a plate, this sampling strategy avoids transition regions between distinct subspaces in the latent space, making it possible to map the sampled points to valid placements.

\textbf{Iterative rescoring.}~Applied to our approach, neighboring values of the latent space tend to be decoded into similar placements.
Therefore, to sample diverse and likely placements, we iteratively re-score the latent distribution $\tilde{p}(z|\mathbf{o})$ around the previous sample.
In iteration $i$, we compute the latent distribution $\tilde{p}_i(z|\mathbf{x}, \mathbf{o}),$ which is normalized using softmax into probability distribution $p(z|\mathbf{x}, \mathbf{o}).$
Latent encoding $z_i$ is sampled from $p(z|\mathbf{x}, \mathbf{o})$.
Then, the latent distribution is re-scored according to:
\begin{equation*}
    \tilde{p}_{i+1}(z|\mathbf{x}, \mathbf{o}) = p_i(z|\mathbf{x}, \mathbf{o}) \cdot \frac{\Vert z - z_i \Vert ^2}{\max\limits_{z} \Vert z - z_i \Vert ^2},
    \label{eq:diversity}
\end{equation*}
which reduces the probability of sampling neighboring embeddings.
This process is repeated in the next iteration.

\subsection{Dataset}
In our framework, all objects are considered rigid and move solely under rigid transformations.
This allows us to interpret any multi-object rearrangement task as a sequence of relative placements between the object being moved and the surrounding scene.
Moreover, given a point cloud scene, we can map any action on the scene with the pair $\langle \mathbf{x}_j, \mathbf{T}_j \rangle$, where transformation $\mathbf{T}_j$ is applied to object $\mathbf{x}_j$.

We train our placement suggester over demonstrations of each task.
The demonstrations are given in the form of a sequence of RGB-D images recorded from two cameras together with the semantic names of the objects in the scene.
The initial observation is converted into a segmented point cloud using the extrinsic camera values and Grounded Segment Anything~\cite{ren2024grounded}.

With two cameras, we can only get partial observations of the scene, which can lead to challenges, for example losing track of the positions of objects when they move to occluded parts of the scene.
To avoid this, we collect all the subsequent data for training by transforming each object's initial point cloud by the transformation applied at that time step in the demonstration. This maintains 3D point correspondence across the transitions of the demonstration.


\subsection{Training Details}
We use a batch size of $4$ and a learning rate of 1e-4.
The size of the input point cloud is $1024$.
In general, we use the same set of standard hyperparameters from TAXPoseD~\cite{wang2024taxposed} to train each of our placement suggesters.
The placement suggesters were trained for $330$k steps, $310$k steps, and $170$k steps for the block stacking, constrained packing, and table bussing tasks, respectively.

\section{Model Deviation Estimator}
\label{app:mde}
The model deviation estimator guides search away from unlikely transitions by predicting the deviation between a planned action in the search tree and its actual execution (Sec.~\ref{sec:mde}).

\subsection{Implementation Details}
To supervise the MDE training, we collect an offline transition dataset from robot-executed plans and label it, similarly to previous work~\cite{lagrassa2022learningmodelpreconditionsplanning}.
Each transition in the dataset consists of a point cloud observation $\mathbf{o}$ and a suggested action $\mathbf{a} = \langle \mathbf{x}, \mathbf{T} \rangle.$
Applying the transformation $\mathbf{T}$ to object $\mathbf{x}$ produces the expected next state $\mathbf{o}'$.

To label this transition with the ground-truth deviation, we execute the action and observe the actual next point cloud $\tilde{\mathbf{o}}.$
Then, we calculate the ground-truth label for this transition $(\mathbf{o}, \langle \mathbf{x}, \mathbf{T} \rangle)$ by computing the sum of object-wise Chamfer distances between the expected next point cloud and the observed next point cloud, divided by the sum of object-wise Chamfer distances between the expected next cloud and the initial point cloud:
\begin{equation}
\sum_{i=1}^{M} \frac{ CD(\mathbf{o}'_{\mathbf{x}_i}, \tilde{\mathbf{o}}_{\mathbf{x}_i}) + \varepsilon}{CD(\mathbf{o}'_{\mathbf{x}_i}, \mathbf{o}_{\mathbf{x}_i}) + \varepsilon},
\label{eq:mde}
\end{equation}
where $i$ indexes each object in the scene, $\mathbf{o}_{\mathbf{x}_i}$ is the point cloud of object $\mathbf{x}_i$, and $CD(\cdot, \cdot)$ is the Chamfer distance.
The numerator measures how different the observed next state is from the expected next state.
The denominator normalizes the deviation by measuring the expected amount of change from the initial state.
To avoid division by zero for static objects, we add a small $\varepsilon.$

If the deviation is high, then it is likely that the next observed state will not match that proposed by the placement suggester, so we want to direct A* search away from such child nodes.

\subsection{Dataset}
We collect an offline transition dataset and label it with ground-truth deviation values for supervised learning of the model deviation estimator (MDE).

We collected $510$ transitions in the simulation block stacking environment.
This was done by repeatedly generating a random initial scene, sampling $k=5$ suggestions for each object from the learned placement suggester, and rolling each of those suggestions out from the initial scene for 1 time step.
We did not train an MDE for the constrained packing tasks.

For the table bussing tasks, we collected $126$ transitions in the real world.
This was done by resetting the scene to some initial state, generating a plan that satisfies the goal using our main method, and then rolling out the plan for as many steps as possible (until either task execution success or failure).

\subsection{Training Details}
We use a PointNet++  network~\cite{qi2017pointnetplusplus} for the MDE architecture.
We use a batch size of $16$ and a learning rate of 1e-3.
We train each MDE for 3000 steps.
The point cloud inputs to the model are mean-centered on the points belonging to movable objects in the scene.

The calculated ground-truth deviations may cover a wide range of values and contain outliers, so we standardize the data by clipping large outliers to a hyperparameter \texttt{clip\_max} and then use MinMaxScaler from scikit-learn to transform the values to the range $[0, 1]$.
See Table~\ref{tab:mde_hyperparams} for specific values of hyperparameters used, including the values used for $\epsilon$ from Equation~\ref{eq:mde}.

\begin{table}[t]
    \centering
    \renewcommand{\arraystretch}{1.2} 
    \scriptsize
    \begin{tabular}{l|cc}
    \toprule
        Hyperparameter & Block Stacking & Table Bussing \\
        \midrule
        \texttt{clip\_max} & 3.2 & 5000  \\
        $\varepsilon$ & 1 & 0.01 \\
    \bottomrule
    \end{tabular}
    \vspace{5px}
    \caption{Hyperparameters used in training the model deviation estimators.}
    \label{tab:mde_hyperparams}
\end{table}

\section{Learned Model Metrics}
\highlight{
We calculated some evaluation metrics on the block stacking task and present the results in Table~\ref{tab:model_metrics}.
We evaluate the object suggester by measuring how often it assigns the ground truth object a probability at least as high as the object would get under a uniform distribution.
This accounts for the fact that multiple objects can be moved at each step.
For the placement suggester, we compute mean translation error $\varepsilon_R$ and rotation error $\varepsilon_t$ in a winner-takes-all fashion using $10$ suggestions, to account for the multimodality of object placements.
For the MDE, we report the train and validation losses.
}
\begin{table}[ht]
    \centering
    \renewcommand{\arraystretch}{1.2}
    \scriptsize
    \begin{tabular}{ccc|cc|cc}
        \toprule
        \multicolumn{3}{c|}{\textbf{Object suggester}} & \multicolumn{2}{c|}{\textbf{Placement suggester}} & \multicolumn{2}{c}{\textbf{MDE}} \\
        Precision & Recall & F1 & $\varepsilon_t$ (m) & $\varepsilon_R$ ($^\circ$) & Train Loss & Val Loss \\
        \midrule
        $0.927$ & $0.926$ & $0.925$ & $0.062$ & $14$ & 3.0e-5 & 1.2e-3 \\
        \bottomrule
    \end{tabular}
    \vspace{5px}
    \captionof{table}{Predictive accuracy of individual learned models.}
    \label{tab:model_metrics}
\vspace{-15pt}
\end{table}

\section{A* Search}
\label{app:nodes}
Our method leverages A* search to search over the space of possible object rearrangements (Sec.~\ref{sec:astar-search}).

\subsection{A* expansion}

Starting from the root node, our method proposes $b = k \cdot M$ child nodes, where $M$ is the number of objects in the scene.
For each object in the scene, a placement suggester suggests $k$ child nodes.
We use $k=10$ for block stacking, $k=5$ for constrained packing, and $k=3$ for table bussing experiments.

After expanding the root node, moving the same object consecutively is equivalent to composing both transformations. Based on this fact and to avoid infinite loops, the last object moved is excluded from generating new suggestions in the next expansion.

\subsection{Cost Function}
\label{app:cost_function}

\textbf{Action cost.}
In both environments, we set the action cost to be 0.01.

\textbf{Collision cost.}
If the action leading to the node is $\langle \mathbf{x}_n, \mathbf{T}_n \rangle$, we reason about collisions of object $\mathbf{x}_n$ with other movable objects in the scene during ``picking" motions and ``placing" motions. Concretely, we voxelize the scene and compute the percentage of voxels belonging to object $\mathbf{x}_n$ that overlap with the other objects in the scene: $C_c(n) = \texttt{voxel\_overlap}(\mathbf{o}_n, \mathbf{x}_n)$

\textbf{Deviation cost.}
This deviation is predicted by a learned model deviation estimator (Sec.~\ref{sec:mde}) as: $C_d(n) = \delta (\mathbf{o}_n, \mathbf{a}_n)$.

\textbf{Probability cost.}
This is derived from the outputs of the learned placement suggester $ p_{\theta}(\mathbf{T} | \mathbf{x}, \mathbf{o})$ and the object suggester $p_{\phi}(\mathbf{x} | \mathbf{o})$.
The likelihood of the action $\mathbf{a}_n = \langle \mathbf{x}_n, \mathbf{T}_n \rangle$ is given by $ p_{\theta}(\mathbf{T} = \mathbf{T}_n | \mathbf{x}_n, \mathbf{o}_{n_{p}})p_{\phi}(\mathbf{x} = \mathbf{x}_{n} | \mathbf{o}_{n_{p}})$, where $\mathbf{o}_{n_{p}}$ is the point cloud of the parent node $n_p$. Therefore, we discourage lower likelihood actions by defining:
$$C_p(n) = 1 - p_{\theta}(\mathbf{T} = \mathbf{T}_n | \mathbf{x}_n, \mathbf{o}_{n_{p}})p_{\phi}(\mathbf{x} = \mathbf{x}_{n} | \mathbf{o}_{n_{p}}),$$
where $w_c, w_d$ and $w_p$ are the weights assigned to the collision, deviation, and probability costs, respectively.

\highlight{
Note that the cost function is completely task-agnostic, consisting of the terms: (1) constant action cost, (2) a voxel-based collision cost, (3) deviation predicted by the MDE, or (4) probabilities output by the learned suggesters.
}

\subsection{Heuristic and Goal Functions}
\subsubsection{Heuristic Functions}
In each of the environments, we calculate the heuristic $h(n)$ directly from the point cloud observation corresponding to node $n$, without access to ground-truth information such as object poses.
Instead, we use the point cloud and object segmentation to estimate the object poses.

In the block stacking environment, the heuristic function is a discrete value indicating how many of the blocks are ``out of place" relative to the goal configuration specified by the goal function.
For example, say the goal is to stack the blocks in red-green-blue order from top-to-bottom, and all three blocks are unstacked on the table.
Then the heuristic value for this state is $2$, since both the green and blue blocks are not in place.
If the robot then stacks the green block on top of the blue block, then the heuristic value decreases to $1,$ since now only the red block is out of place.
To determine whether the blocks are in the desired relative positions to satisfy the goal function, we estimate the pose of each block from its point cloud and then check their heights and whether their $(x, y)$ coordinates are aligned, within a certain error threshold. 

In the table bussing environment, the heuristic function is a continuous value indicating how far the objects are from a target point.
The target point is selected as the center of either one of the plates' point clouds.
Each object in the scene also has a specified center, calculated as:
$$c = \frac{\max(x, y) + \min(x, y)}{2}$$
where $x, y$ are the coordinates of the object's point cloud in the world frame.
The heuristic for node $n$ is therefore defined as the sum of the Euclidean distances of each object's center to the target point on the $XY$ plane, as:
$$h(n) = \sum_{i=1}^{M} c_i(x, y)$$
where $M$ is the number of objects in the scene.

We ensure that the heuristic function underestimates the distance from the current node to the goal, so any goal state will have heuristic value $0$.

\highlight{
Note that only the heuristic function $h(n)$ needs to be designed for a new task.
The heuristic function in search-based methods does not need to be heavily engineered because it is only meant to serve as a rough guide; the search can backtrack from dead ends and continue onto alternative paths (see Appendix~\ref{app:search_graphs} for search graph visualizations).
}
Since heuristics are not the main focus of this paper, we leave the task of learning these heuristic functions from demonstration data to future work.

\subsubsection{Goal Functions}
In the block stacking environment, the goal function is closely related to the A* heuristic function; namely, a node $n$ whose corresponding point cloud observation $\mathbf{o}$ satisfies the goal function $\mathcal{G}(\mathbf{o}) = 1$ if and only if it has $h(n) = 0.$

In the table bussing environment, the goal function checks if all of the objects are stacked on top of a plate. This is done by checking if the Euclidean distance in the $XY$ plane from each object's center to the reference plate's center is below a certain threshold.

\highlight{
Similar to the heuristic function, the goal function needs to be designed for a new task.
Since goal functions are not the main focus of this paper, we leave the task of learning them from demonstration data to future work.
}

\subsection{Multi-goal A*}
\label{sec:multigoal}
Rather than terminating search immediately upon finding a state that satisfies the goal condition, we let search continue until one of the following occurs:
1) the open list is empty, 2) the node expansion limit is reached, or 3) $m$ goals have been found, where $m$ is a hyperparameter.
During search, we record all goal nodes: nodes that satisfy the goal condition $\mathcal{G}(\mathbf{o}_n) = 1$.
After search terminates, we select one of the goal nodes using a score function and output the corresponding path.

Once selected, we backtrack from the goal node $n_g$ to the initial node $n_1$, which gives us a sequence of actions $\mathbf{a}$ that constitute the output plan $P = \{\mathbf{a}_1, \mathbf{a}_2, \dots \mathbf{a}_{T-1}\}$. We then attempt to execute this plan with a robot manipulator.
If A* search exceeds a maximum node expansion budget, we say that no plan is found and terminate the search.

This outputs multiple possible goal nodes, from which we select one using a score function and output its corresponding path.
In real-world table bussing, we use $m=10$, and the score function selects the path with the lowest sum of collision costs.
In simulation, we use $m=1$, and the path whose goal node has the best block stacking alignment is chosen.

\section{Experimental Setup}
\label{app:experimental_setup}

\subsection{Datasets and Tasks}

\begin{figure}[t]
\centering
    \begin{minipage}{\linewidth}
        \centering
        \includegraphics[width=0.8\linewidth]{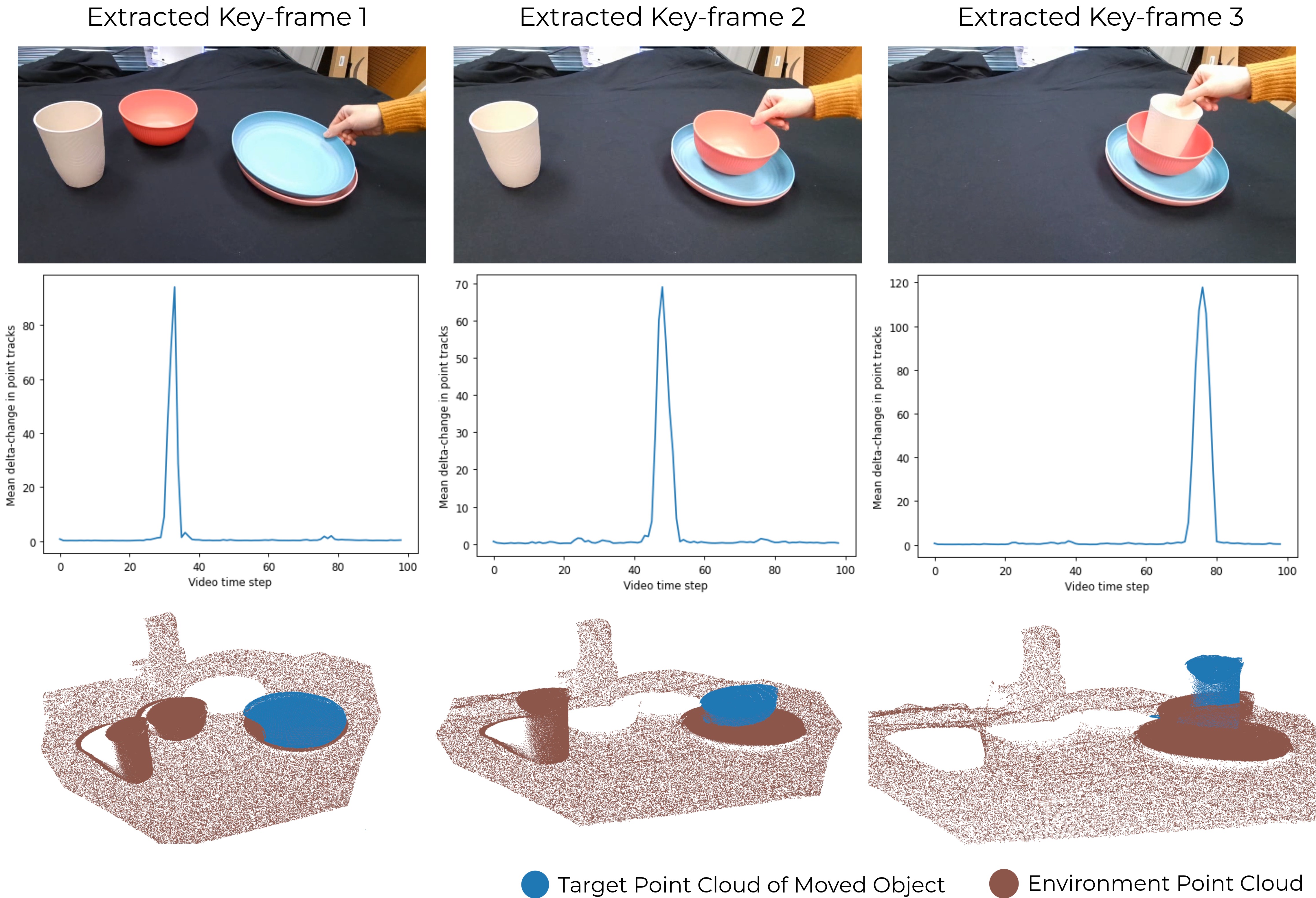}
        \captionof{figure}{\textbf{Real World Data Collection}. Our demonstration data consists of RGB-D videos of a human moving the objects in the scene.
We use Grounded Segment Anything~\cite{ren2024grounded} and CoTracker3~\cite{karaev2024cotracker3} to extract 2D tracks from which we extract a 2D velocity-time graph for each object. For each detected object movement, RANSAC-based SVD computes the best rigid transformation (4×4 matrices) representing that movement.}
        \label{fig:data_collection}
    \end{minipage}
    \vspace{-8px}
\end{figure}

\textbf{Block Stacking (Simulation).}
Our demonstration data for this task consists of 96 transitions $(\mathbf{o}_j, \mathbf{a}_j)$ collected using two depth cameras in PyBullet~\cite{coumans2020}. We collect data in simulation by using the keyboard to control the pose of the object. The user chooses which object to move and then controls the pose of the object until it reaches the desired goal pose. Then, we record the final transformation that the object encountered after the user is done interacting with it. We store the point cloud observation from simulation, the object moved, and the extracted transformation in the format $(\mathbf{o}_j, \mathbf{a}_j)$.
We train and test on a version of the environment with three blocks.

\textbf{Constrained Packing (Simulation).}
Our demonstration data consists of $8000$ transitions from a dataset containing 4-step task demonstrations.
This dataset was released by the authors of Points2Plans~\cite{huang2024points2plans}.

\begin{figure}
    \centering
    \includegraphics[scale = 0.12]{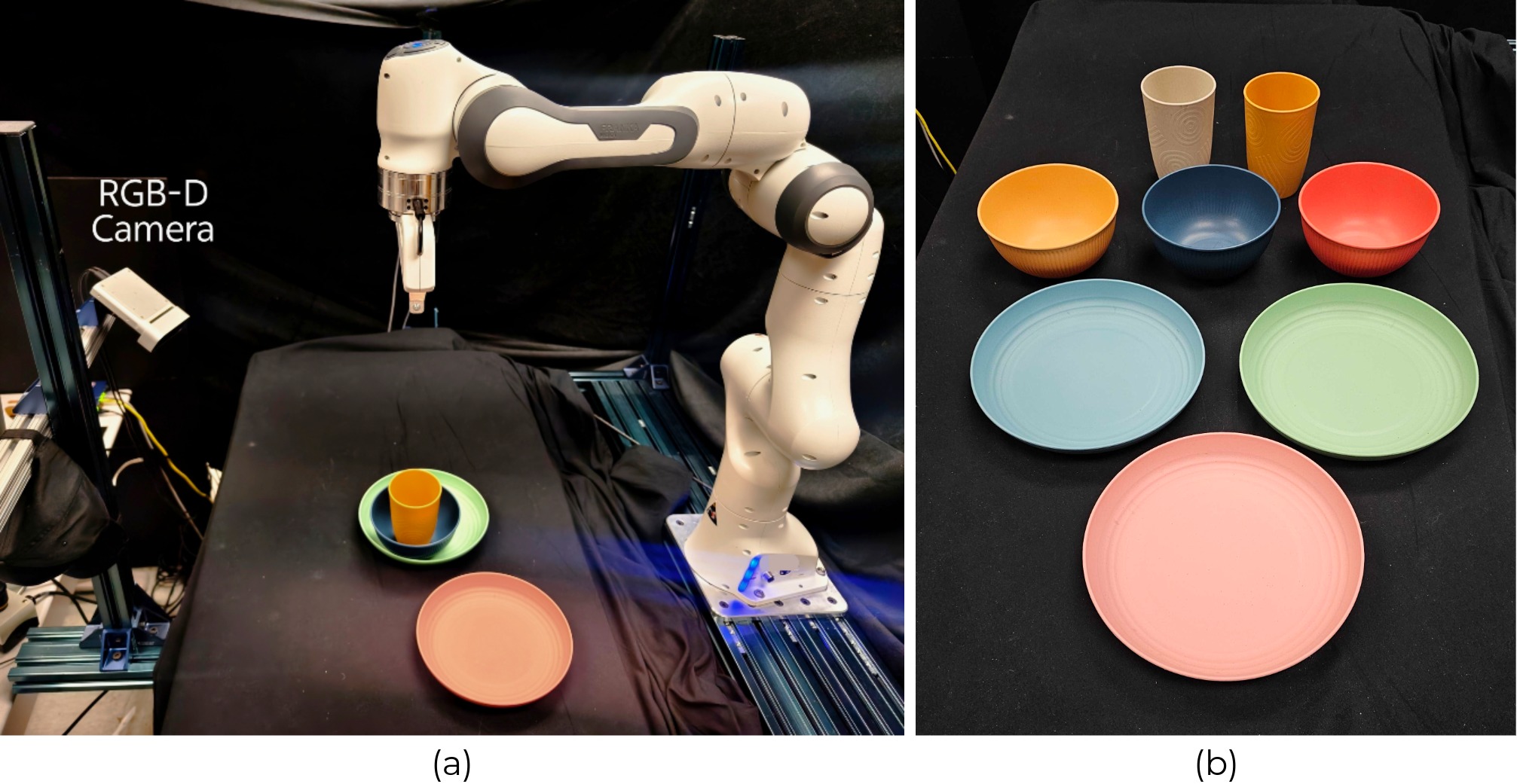}
    \caption{\textbf{Real-world setup.}
    Left: Our real-world setup of the table bussing environment with the RGB-D camera, Franka arm, and a set of objects. Right: All of the objects (plates, bowls, and cups) seen in the table bussing environment.}
    \label{fig:hardware}
\end{figure}

\textbf{Table Bussing (Real World).}
Our demonstration data for this task consists of 156 transitions collected using an RGB-D video camera in the real world, by a human moving the objects in the scene (Figure~\ref{fig:hardware}).
We use Grounded Segment Anything~\cite{ren2024grounded} to segment out each object in the video. Then, using CoTracker3~\cite{karaev2024cotracker3}, we extract 2D tracks across video frames for each object in the scene. By applying a threshold to the 2D tracks, we identify stationary periods and movement periods for each object in the scene. Assuming that every object moves independently of the other objects, we determine the order of object movements. For each detected movement, RANSAC-based SVD computes the best rigid transformation (4×4 matrices) between stationary states of each object (See Figure~\ref{fig:data_collection}). The output is a labeled initial point cloud and a sequence of transformations describing how objects were manipulated, stored in the format $(\mathbf{o}_j, \mathbf{a}_j)$.
We evaluate our method on two versions of this environment: 1) two plates, a bowl, and a cup, 2) one plate, two bowls, and a cup.

\subsection{Evaluation Details}
Simulation block stack experiment results are based on a suite of 23 different initial scene configurations, with runs over five different seeds.
The evaluation results are averaged and show a 95\% confidence interval.
Real-world table bussing experiment results are based on a suite of 14 different initial scene configurations. These configurations range in complexity from 2 to 5 steps, where a step refers to a single object reconfiguration. Figure~\ref{fig:configurations} illustrates examples of 4-step, 3-step, and 2-step configurations in both simulation and real world settings.

For real-world experiments, we use Contact-GraspNet~\cite{sundermeyer2021contact} to find appropriate grasps and then execute the transformations with pick-and-place motion primitives.
These execution methods can find a robot path based on the current point observation and the predicted action, without any extra ground-truth information.

\subsection{Points2Plans Comparison}
\textbf{Constrained Packing.}
We define success in the constrained packing task to be when all of the objects are arranged on top of the shelf surface without being stacked on top of each other.
This is a modified but similar version of the success metric used in Points2Plans~\cite{huang2024points2plans}, since their metric depends on goal predicates and feasibility-related predicates -- for example, a predicate that determines when an object’s placement location is blocked by another object that has already been placed.
Our goal function cannot mimic theirs exactly since we do not use relational abstractions; as a result, ours differs in that it permits placing objects behind other objects, as long as the already-placed objects are not disturbed.

\textbf{Evaluation.}
The authors of Points2Plans~\cite{huang2024points2plans} have only released 5 example scenes from their test dataset, so we run each of the scenes with 100 different random seeds and report the success rates for these 500 resulting runs in Figure~\ref{fig:points2plans_comparison}.

We also modify the action primitives from Points2Plans~\cite{huang2024points2plans} to better align with our method.
Specifically, rather than using MoveIt’s inverse kinematics, we compute the Jacobian of the delta pose between the current pose and a pre-defined grasp pose of a block.
We then use this Jacobian to apply joint velocities to the manipulator.
This process is repeated inside a control loop until the robot reaches the grasp pose.
We leverage this control loop to try each available grasp pose until the block is grasped successfully.
Since grasping is orthogonal to our contribution, we employ a model similar to a suction gripper: when the manipulator reaches within a delta error of the block’s grasp pose, we attach the block to the gripper and immediately close the fingers to ensure a stable grasp.

\subsection{3D Diffusion Policy Baseline}
\label{app:dp3_baseline}
In Table~\ref{tab:results_baseline}, we compare the execution performance of our method with 3D Diffusion Policy (DP3)~\cite{Ze2024DP3} as a baseline.

\textbf{Dataset.} We train DP3 for a block stacking task, where the goal is to stack the blocks in red, green, blue order from top to bottom.
There is only one goal configuration in the demonstrations since DP3 is not goal-conditioned.
The training dataset has 23 initial states where the red, green, and blue blocks are arranged in various positions, either stacked or unstacked in different configurations.
We collected 23 expert demonstrations in the PyBullet simulator~\cite{coumans2020}, utilizing Pybullet's inverse kinematics tool and collision checker.
For each movement of a block, 240 observations and state-action pairs are collected.
The observations are the dense, segmented point clouds of the blocks.
The states and actions consist of the end effector's  x, y, z coordinates, quaternion, and the gripper widths.

\textbf{Training.} The DP3 agent was trained with an action execution horizon of 4 and an observation history of 2.
All training parameters are the same as those in the original DP3 model~\cite{Ze2024DP3}.
\highlight{
DP3 has 280M parameters, compared to 26.7M parameters of our three learned models in combination.
}

\textbf{Evaluation.}
Since DP3 is not goal-conditioned, the goal configuration at evaluation time is consistent with the one seen in the training dataset.
For a fair comparison with our approach, we thus evaluate both DP3 and our method on five different tasks consisting of five different initial blocks configurations with the same goal.

The evaluation metrics are the success rate and the task completion rate.
For each initial configuration in the test dataset, success rate is the total number of times the DP3 agent successfully stacks all three blocks in the red-green-blue order divided by the total number of trials.
For each initial configuration, 2 to 4 steps are required to complete the task.
Each step involves stacking or unstacking a block.
The task completion rate is the average number of steps completed divided by the total number of steps required for an initial configuration.

\subsection{Random Rollouts}
\label{app:random_rollouts}
One might wonder whether the observed improvement with the search method merely results from the number of evaluated nodes.
To investigate this, we replace A* search with a series of random rollouts.
This method chooses a child node uniformly at random from $b=k\cdot M$ possible child nodes, where $M$ is the number of objects.
Compared to Beam Search, this method should see more diverse expansions.

In this ablation study, we allow the Random Rollouts method to continue rolling out action trajectories until the total number of expanded nodes across all of the rollouts exceeds the node expansion budget of our main method.
Each rollout is allowed to continue to a maximum path length (6 steps).
For block stacking tasks, we allow the search to expand a maximum of $200$ nodes.
We choose to normalize random rollouts by the total number of node expansions since we did not optimize our code in terms of wall-clock runtime -- we leave this to future work.
As a reference, our method takes between 1 and 113 seconds to plan for a single task, with an average planning time of 27 seconds per task.

\section{Experimental Results}
\label{app:experimental_results}



\subsection{Task Execution Success Rate vs. Task Complexity}
\begin{figure}
    \centering
    \includegraphics[width=0.47\textwidth]{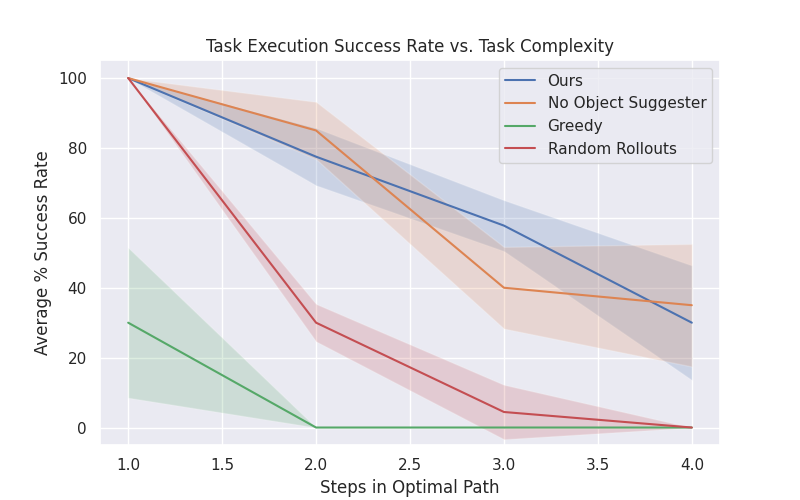}
    \caption{Task execution success rate as a function of task complexity in the simulation block stacking environment. Results are averaged over 5 seeds and show a 95\% confidence interval.}
    \label{fig:sim_execution}
\end{figure}
\begin{figure}
    \centering
    \includegraphics[width=0.47\textwidth]{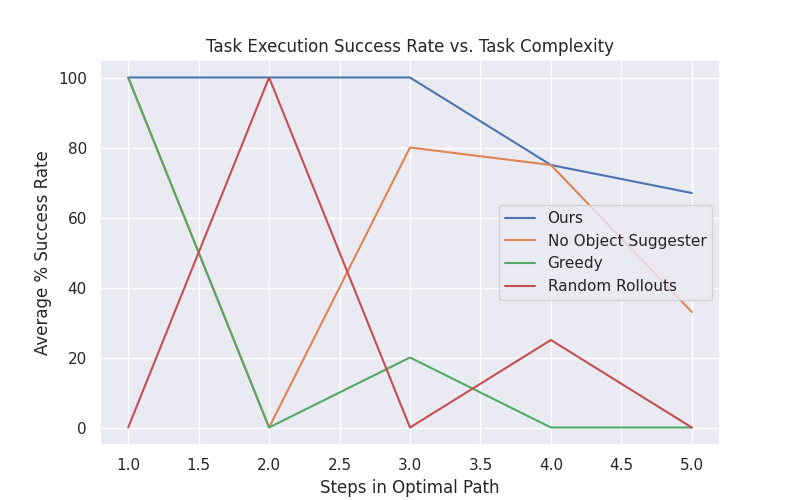}
    \caption{Task execution success rate as a function of task complexity in the real-world table bussing environment.}
    \label{fig:real_execution}
\end{figure}

Figures~\ref{fig:sim_execution} and \ref{fig:real_execution} show task execution success rates of our method and ablations as a function of task complexity in simulation and in the real world, respectively.
These reinforce the conclusions that:
\begin{itemize}
    \item The Greedy ablation is too myopic to achieve much success on tasks that require more than one step to a goal configuration.

    \item The Random Rollouts ablation achieves some success, but its performance rapidly deteriorates as task complexity increases beyond two steps.

    \item Our main method outperforms the ablation with no object suggester.
\end{itemize}

\subsection{Qualitative Analysis: Search Graphs}
\label{app:search_graphs}
\begin{figure*}[h!]
    \centering
    \includegraphics[width=0.4\textwidth]{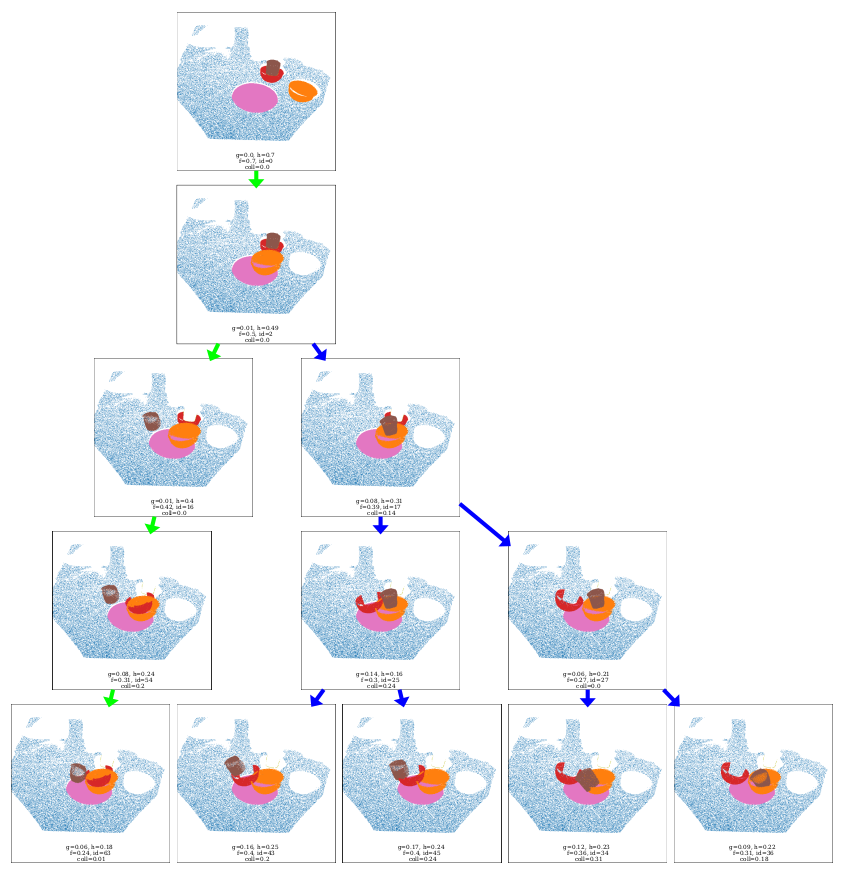}
    \caption{\textbf{Expanded graph example 1}. The graph represents the expanded nodes for a 2-step table bussing configuration where a bowl is on a plate and a cup is inside a bowl on the table. A plan is found that first moves the cup to the table, then stacks the bowls, and places the cup on the plate. The plan found is marked with green colored edges }
    \label{fig:eg1}
\end{figure*}

\begin{figure*}[h!]
    \centering
    \includegraphics[width=0.8\textwidth]{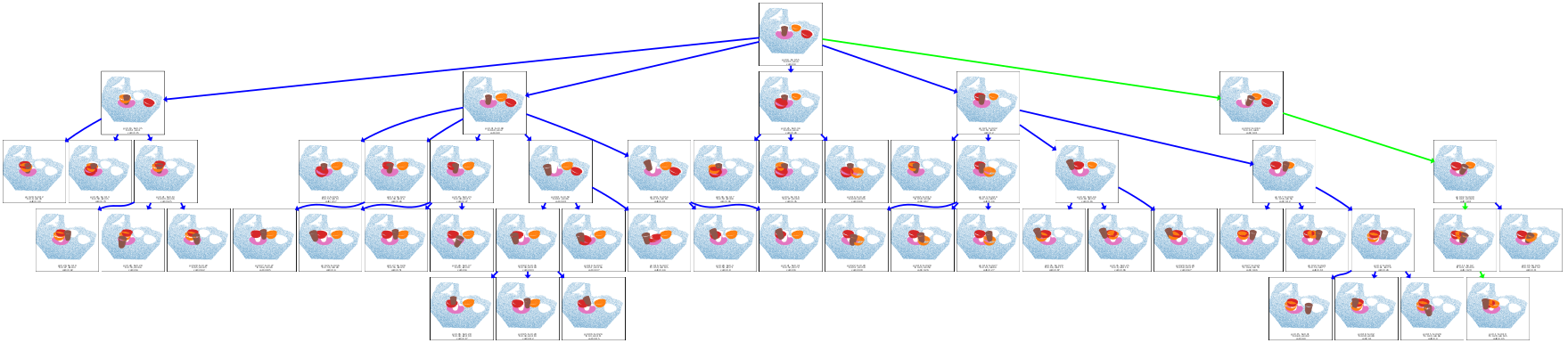}
    \caption{\textbf{Expanded graph example 2}. The graph represents the expanded nodes for a 4-step table bussing configuration where a cup is initially placed on a plate. A plan is found that first moves the cup to the table, then stacks both the bowls, and then moves the cup back. The plan found is marked with green colored edges}
    \label{fig:eg2}
\end{figure*}

\begin{figure*}[h!]
    \centering
    \includegraphics[width=0.8\textwidth]{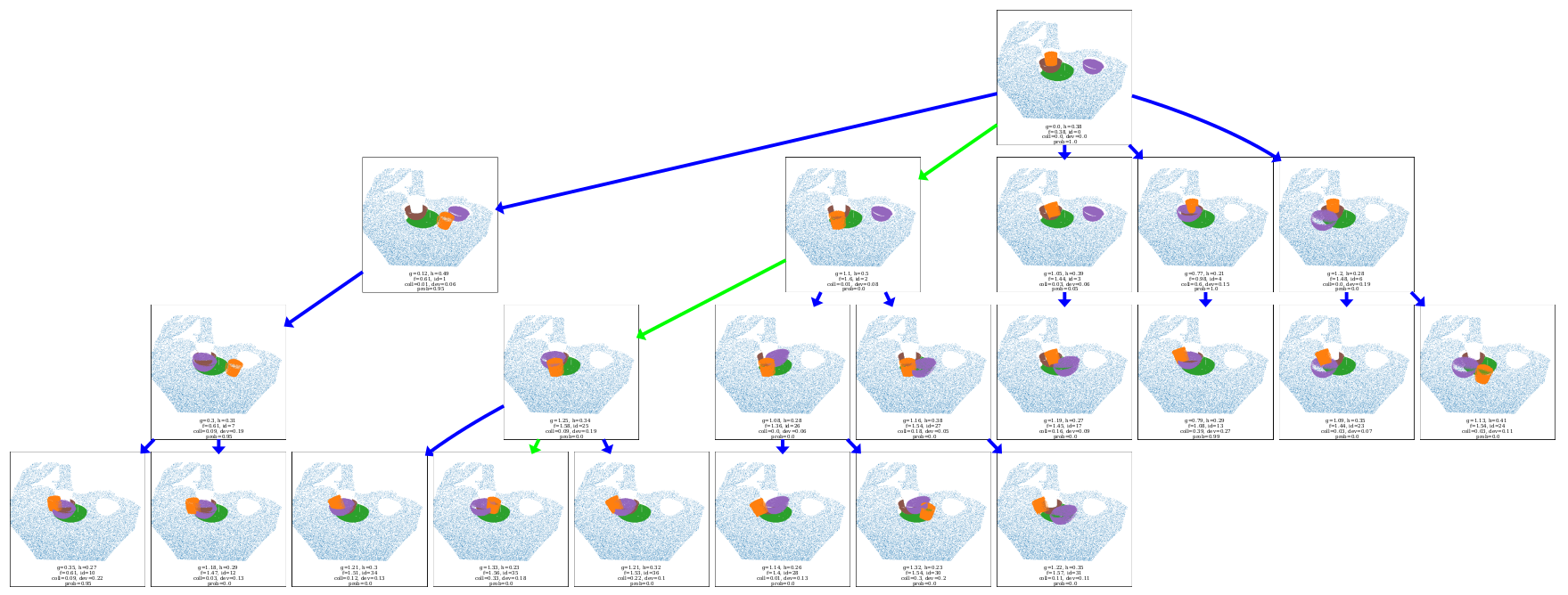}
    \caption{\textbf{Expanded graph example 3}. The graph represents the expanded nodes for a 3-step table bussing configuration where a cup is initially placed inside a bowl on top of the plate, and a separate bowl is placed on the table. A plan is found that first moves the cup to the table, stacks the remaining bowl from the table on top of the bowl already on the plate, then moves the cup back inside the bowls. The plan found is marked with green colored edges}
    \label{fig:eg3}
\end{figure*}

We visualize 2 types of graphs for our method:

1) An expanded graph that marks all the nodes expanded during search. Plans found inside this graph are marked via green edges.
Figures~\ref{fig:eg1} through~\ref{fig:eg3} show expanded graphs for three different table bussing initial configurations.

\begin{figure*}[h!]
    \centering
    \includegraphics[width=0.8\textwidth]{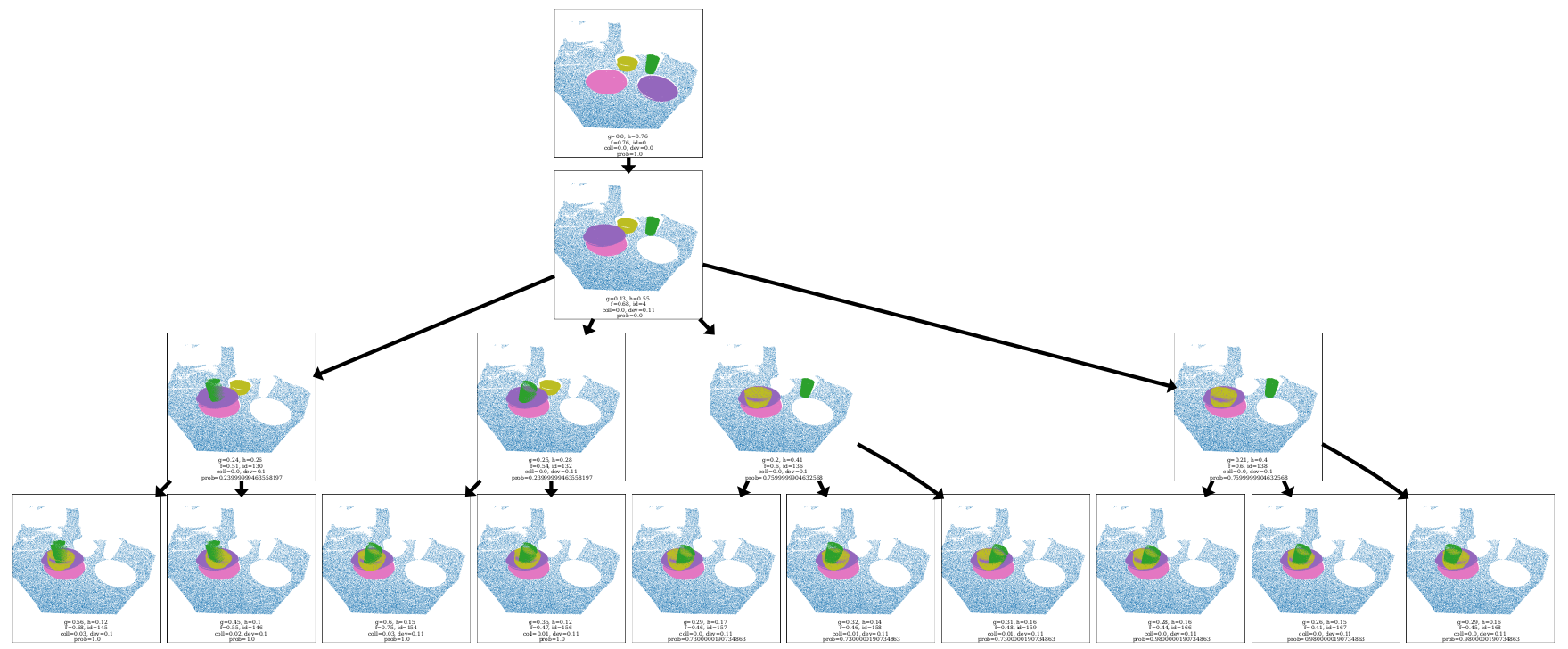}
    \caption{\textbf{Plan graph example 1}. The graph represents all of the plans and goals found for a 3-step table bussing configuration with multi-goal A* where 2 plates, a bowl and a cup are placed separately on the table. We see in this graph that A* search finds \textbf{multimodal} paths to achieve the goal. It may either produce a plan that places the cup on the plate first and then stacks the bowls, or it may stack the bowls first and then place the cup inside the bowls.}
    \label{fig:pg1}
\end{figure*}

\begin{figure*}[h!]
    \centering
    \includegraphics[width=0.8\textwidth]{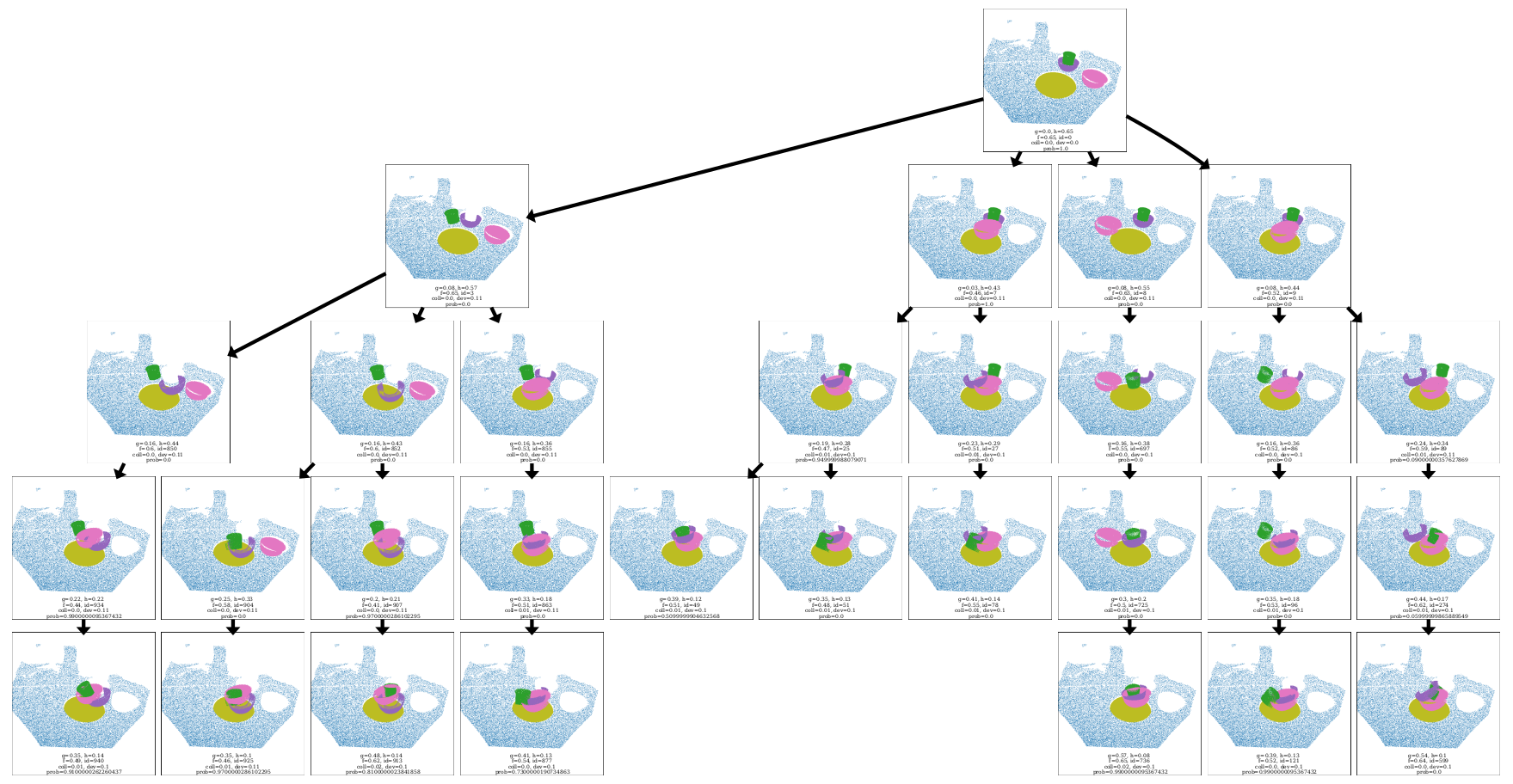}
    \caption{\textbf{Plan graph example 2}. The graph represents all of the plans and goals found for a 3-step table bussing configuration where a plate is placed on the table, alongside a cup inside a bowl, with another bowl placed separately on the table. We see in this graph that A* search finds \textbf{multimodal} paths to achieve the goal. It may either move the cup inside the bowl to the table first, and then stack the bowls before placing the cup back inside. Or, it may first move a bowl on top of the plate, move the cup to the table, and stack the bowls before placing the cup back inside. Illegal plans that move the bowl when a cup is on top of it are not chosen due to their high collision score.}
    \label{fig:pg2}
\end{figure*}

\begin{figure*}[h!]
    \centering
    \includegraphics[width=0.8\textwidth]{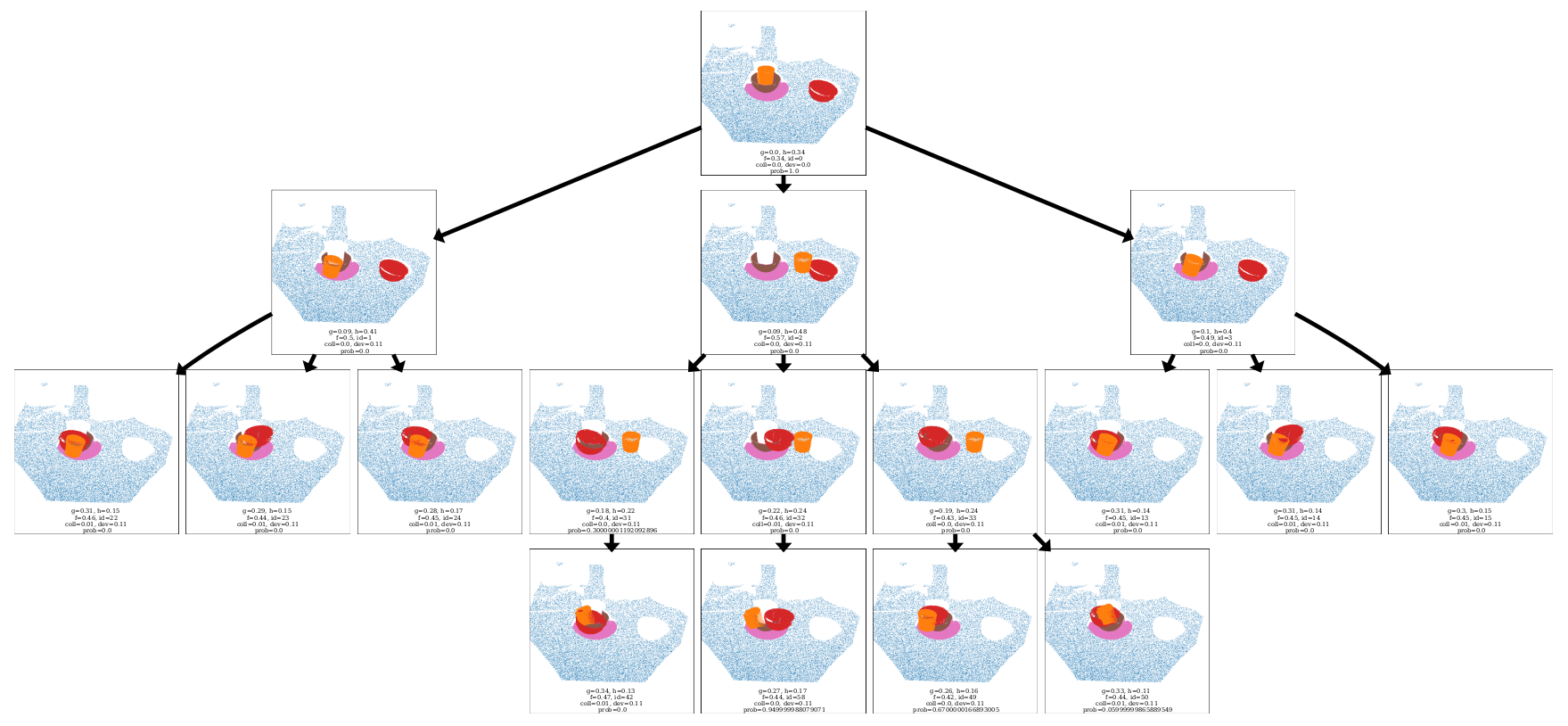}
    \caption{\textbf{Plan graph example 3}. The graph represents all of the plans and goals found for a 3-step table bussing configuration where a cup is placed inside a bowl on a plate, alongside another bowl placed separately on the table. We see in this graph that A* search finds \textbf{multimodal} paths of \textbf{varying lengths} to achieve the goal. The cup may be first moved to the table, then the bowls can be stacked, before moving the cup back inside the bowls. Or, the cup can be precisely moved to the side of the plate, which leaves an opening for the bowls to be stacked directly. The former plan achieves the goal in 3 steps, while the latter achieves the goal in just 2 steps. Illegal plans with high collision are not selected.}
    \label{fig:pg3}
\end{figure*}

2) A plan graph that visualizes different paths to goal configurations found during multi-goal A* search.
Figures~\ref{fig:pg1} through~\ref{fig:pg3} show plan graphs for three different table bussing initial configurations.



\begin{figure*}[h!]
    \centering
    \includegraphics[width=0.2\textwidth]{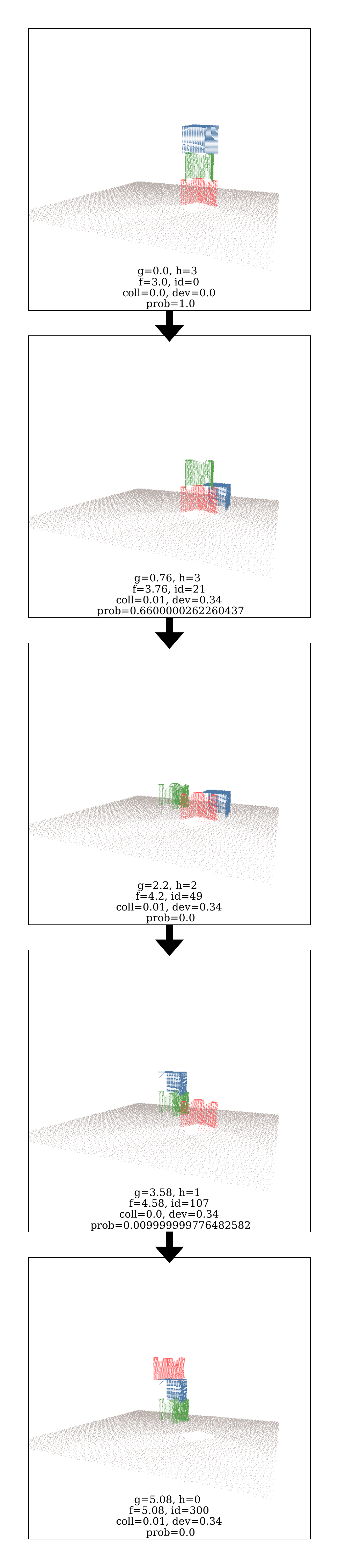}
    \caption{\textbf{Plan graph example 4}. This graphs shows a plan found for a 4-step block stacking initial configuration. Since we set $m=1$ for multi-goal search, the graph contains only one plan, which is the first plan that was found during search.}
    \label{fig:pg4}
\end{figure*}


\end{document}